\documentclass[letterpaper,10pt,conference]{ieeeconf}

\IEEEoverridecommandlockouts
\usepackage{cite}

\usepackage[utf8]{inputenc}
\usepackage[T1]{fontenc}
\usepackage{url}
\usepackage{booktabs}
\usepackage{amsmath,amssymb,amsfonts}
\usepackage{graphicx}
\usepackage{multicol,multirow}
\usepackage{makecell}
\usepackage{array}
\usepackage{float}
\usepackage{pifont}
\usepackage[table,dvipsnames]{xcolor}
\usepackage[bookmarks=false]{hyperref}

\newcommand{\Method}{{Praxis}}

\newcommand{\cmark}{\textcolor{green!80!black}{\ding{51}}}
\newcommand{\xmark}{\textcolor{red!90!black}{\ding{55}}}

\title{\LARGE\bf \Method{}: Distilling Physical Interaction Priors from Egocentric Videos for Generalizable Whole-Body Manipulation}

\author{}

\makeatletter
\def\ps@hypotheticaldraft{%
\def\@oddhead{}\def\@evenhead{}%
\def\@oddfoot{\hfil\footnotesize INTERNAL DRAFT -- OKAMI VALUES ARE HYPOTHETICAL, NOT MEASURED\hfil}%
\let\@evenfoot\@oddfoot}
\makeatother

\input{teaser}
\usepackage{fontawesome5}
\usepackage{hyperref}
\DeclareRobustCommand{\AuthorNote}{%
    \parbox[t]{0.95\linewidth}{%
        \raggedright
        ${}^{*}$Equal contributions.
        ${}^{\dagger}$Corresponding author.
        ${}^{1}$School of Data Science, The Chinese
        University of Hong Kong, Shenzhen.
        ${}^{2}$Sun Yat-sen University.
        ${}^{3}$JD Technology.
        Email: \texttt{liuguiliang@cuhk.edu.cn}%
    }%
}
\author{
    Shuliang He${}^{1*}$,
    Ruiyan Xu${}^{1*}$,
    Bo Yue${}^{1*}$,
    Hengming Zhang${}^{1}$,\\
    Huayi Zhou${}^{1}$,
    Shuai Wang${}^{3}$,
    Wei-Shi Zheng${}^{2}$,
    Guiliang Liu${}^{1\dagger}$\\\\
    \faGlobe\;Project Website: \href{https://edem-ai.github.io/Praxis/}{https://edem-ai.github.io/Praxis/}
    \thanks{\AuthorNote}
}
\begin{document}
\bstctlcite{BSTcontrol}
\raggedbottom \setlength{\textfloatsep}{8pt} \setlength{\dbltextfloatsep}{12pt plus 2pt minus 2pt} \setlength{\intextsep}{10pt plus 2pt minus 2pt}

\maketitle
\thispagestyle{empty}
\pagestyle{empty}

\begin{abstract}
Mobile humanoid manipulation requires both reaching a usable workspace and preserving precise hand–object interactions as object poses and contact conditions change. Learning these behaviors from limited task-specific data remains challenging.
To bridge this gap, we introduce Praxis, a whole-body manipulation framework that combines physical interaction priors from one-shot egocentric video demonstrations with closed-loop posture calibration and online perception. The framework coordinates three stages: vision-language-guided navigation toward target objects, closed-loop posture calibration to align the arm-hand workspace, and dexterous manipulation with synchronized upper- and lower-body control. Online visual feedback re-grounds demonstrated interaction geometry under new object poses and scene configurations, while tactile feedback adapts hand motions to actual contact conditions. 
Each manipulation skill is specified by one human demonstration, without task-specific manipulation-policy retraining.
Experiments across five long-horizon manipulation tasks demonstrate spatial, visual, and cross-object generalization, as well as recovery from external physical disturbances across all three stages.
\end{abstract}

\begin{figure*}[!t]
    \centering
    \includegraphics[width=\linewidth,pagebox=cropbox]{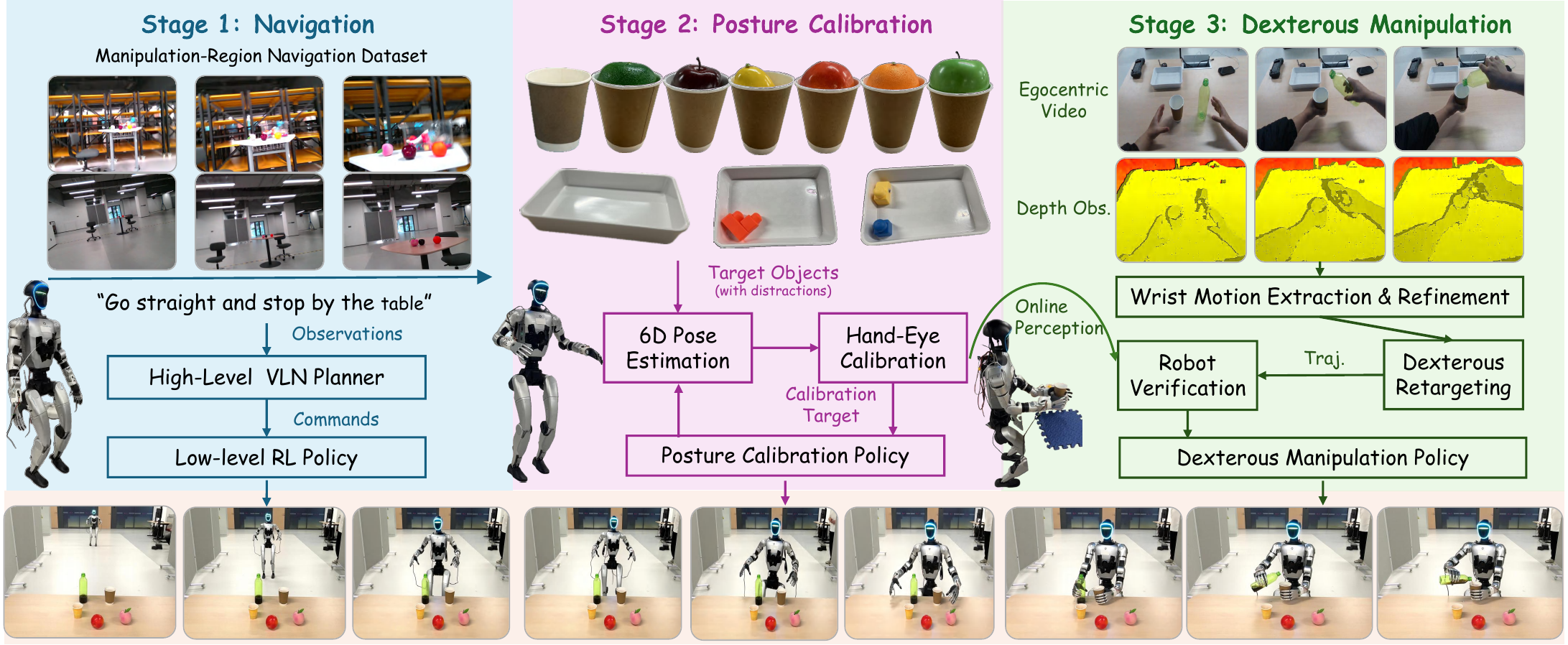}
    \caption{\Method{} architecture. Stage 1: Navigation to the manipulation region. Stage 2: Calibration to a manipulation-ready posture in a closed-loop manner. Stage 3: Distilling hand-object interaction priors through egocentric videos and applying online feedback to adapt manipulation policies.}
    \label{fig:pipeline-detailed}
\end{figure*}

\section{Introduction}
Developing general-purpose robotic manipulation policies is a foundational goal of embodied intelligence. Recent systems generalize across objects, environments, tasks, and embodiments~\cite{intelligence2025pi,bjorck2025gr00t,team2025gemini}, but their success remains largely confined to static workspaces of fixed-base robots. Extending dexterity to larger, dynamically changing environments from limited task-specific data remains challenging.

Humanoid robots can expand the reachable workspace through mobile bodies, articulated torsos, high-DoF arms, and dexterous hands. The development of Whole-Body Control (WBC) algorithms offers a pathway to coordinate these capabilities under a unified control framework~\cite{gu2025humanoid}. Learning-based methods train policies with reinforcement learning (RL) in simulation and transfer them to real robots~\cite{liu2025trajbooster,weng2025hdmi}. Although they produce expressive dancing and athletic motions~\cite{xie2025kungfubot}, extending these approaches to dexterous manipulation, which requires rich contact and precise physical interaction, remains an open problem.

We argue that whole-body manipulation requires a system that brings objects into a reachable arm-hand workspace while preserving task-critical hand-object interactions under changing object poses and contact conditions.
Drawing inspiration from the emergence of human-to-robot transfer~\cite{kareer2025emergence}, we introduce \textbf{\Method{}}, a hierarchical framework that couples interaction priors from a single egocentric RGB-D video with closed-loop posture and perception adaptation.
Such demonstrations supply dense semantic and motion cues for dexterous interaction, reducing dependence on labor-intensive robot data. Our key insight is to preserve demonstrated object-relative interaction geometry while adapting body posture and contact-dependent motions online.

As shown in Fig. ~\ref{fig:pipeline-detailed}, 
Praxis coordinates three stages:
(1) vision-language navigation to the target vicinity;
(2) closed-loop posture calibration to align the arm-hand workspace with the objects; and
(3) manipulation using demonstrated keyposes while maintaining whole-body balance.
Online visual and tactile feedback re-grounds object-centric interaction priors, adapting execution to new object poses, scene layouts, and contact conditions using one demonstration per task without task-specific manipulation policy retraining.

We evaluate \Method{} on five long-horizon whole-body manipulation tasks, with additional long-distance evaluations on \texttt{Hand Over} and \texttt{Pour Water}.
Under environmental variations, \Method{} achieves an average success rate of $76.97\%$, compared with $66.67\%$ for YOTO, the strongest baseline in this evaluation.
Long-distance success rates average $75\%$ at $3$\,m and $50\%$ at $5$\,m across both tasks and approach protocols.
The largest task-level margin over YOTO occurs on Manipulate Pipette, with success rates of $63.64\%$ and $45.45\%$, respectively.



Overall, our main contributions are:
\begin{itemize}
    \setlength{\itemsep}{0pt}
    \setlength{\parsep}{0pt}
    \item A one-demonstration framework for long-horizon mobile humanoid manipulation. Each manipulation skill is specified by a single egocentric RGB-D demonstration, without task-specific manipulation-policy retraining.
    \item A closed-loop execution scheme that transfers demonstrated interactions through object-centric motion adaptation, posture calibration, and tactile grasp regulation.
    \item Real-world evaluation on five manipulation tasks, including additional long-distance execution, disturbance recovery, and component ablations.
\end{itemize}

\section{Related Work}

\textbf{Humanoid Whole-Body Manipulation.}
Whole-body manipulation couples locomotion, postural stability, and dexterous end-effector control. Frameworks such as WoCoCo~\cite{zhang2024wococo} and FALCON~\cite{zhang2025falcon} use contact primitives or decoupled sub-agents, while other learning-based systems learn unified policies from teleoperated imitation or RL~\cite{fu2024humanplus,he2024omnih2o,yue2025real,yue2025understanding}. Adapting these to new task geometries requires extensive data, retraining, or manual engineering. \Method{} instead extracts interaction priors from a single video and re-grounds them through real-time perception for generalization and robustness.

\textbf{Learning from Human Hand Videos.}
Human hand videos provide a scalable source of supervision for learning complex manipulation behaviors \cite{liu2024taco,grauman2024ego}. To mitigate the embodiment gap between humans and robots, prior work leverages intermediate representations, such as keypoints \cite{gao2024bi}, affordances \cite{li2024learning}, and geometric correspondences \cite{zhang2024one}, to enable motion retargeting \cite{shaw2024learning} and downstream action planning \cite{kerr2024robot}.
The closest works are YOTO and OKAMI~\cite{li2024okami}. YOTO learns bimanual manipulation from binocular videos using stereo depth estimation and fixed-base arms with parallel grippers. OKAMI uses continuous trajectory warping for stationary upper-body control. \Method{} combines RGB-D-based dexterous retargeting with navigation and posture calibration, extracts discrete interaction keyframes, and uses tactile grasp regulation and online recovery for long-horizon execution.

\section{Methodology}\label{main:method}

\begin{figure}[htbp]
    \centering
    \includegraphics[width=0.8\linewidth]{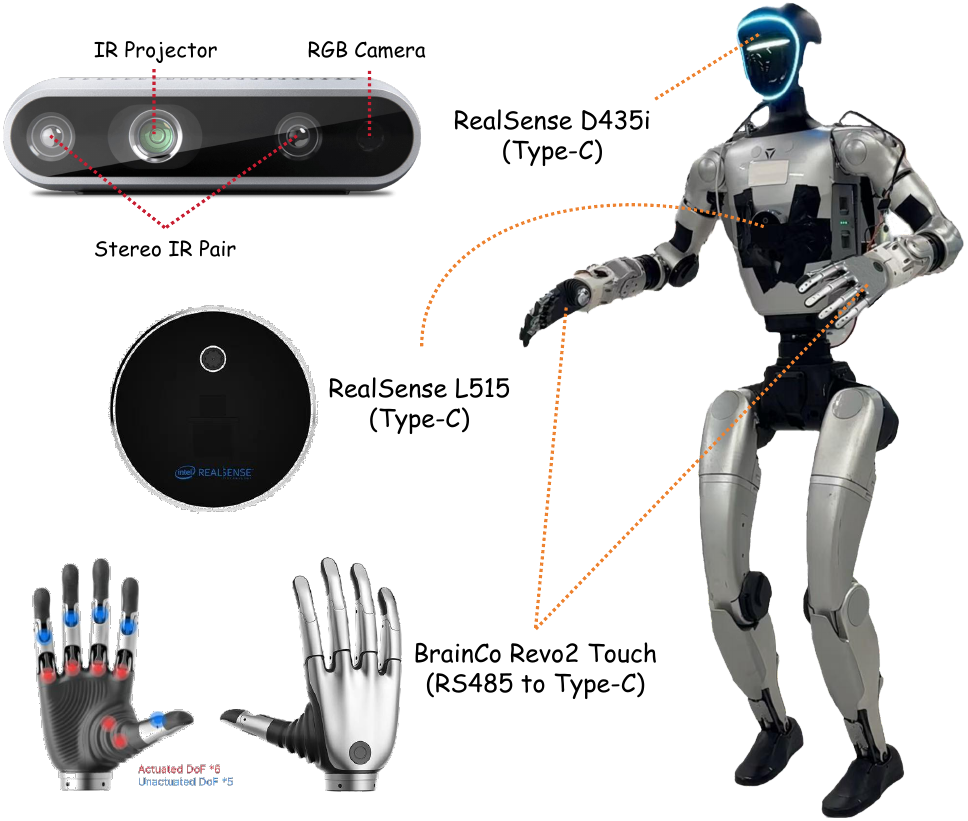}
    \caption{Hardware system of \Method{}.}
    \label{fig:hardware}
\end{figure}

As shown in Fig.~\ref{fig:hardware}, we implement \Method{} on a Unitree G1 with two BrainCo Revo2 hands, each with five fingers, tactile sensing, and $6$ actuated DoFs coupled to $11$ joints. A torso-mounted RealSense L515 supports navigation, and the head-mounted D435i supports calibration and manipulation. Camera streams run at $25$ Hz, while asynchronous I/O supports whole-body control at $50$ Hz.
All three stages share the same whole-body balance controller.

\subsection{Humanoid WBC for Navigation} 

Navigation brings distant objects into the humanoid's workspace. Following generalist navigation models~\cite{Wu2024VLNSurvey}, a high-level VLN planner predicts symbolic commands and a low-level WBC policy executes balanced locomotion.
We initialize the high-level policy $\pi^{\text{up}}$ from the official NaVILA~\cite{cheng2024navila} checkpoint already trained on large-scale navigation data. We adapt it with LoRA on standardized navigation templates that explicitly label safe pre-manipulation stopping regions, as the original checkpoint struggles to respond appropriately to stop commands.

We collect $150$ real-world expert trajectories: $46$ forward trajectories, $32$ per turning direction (left and right), and $20$ per lateral direction (left and right). 
The seven navigation targets (six tables and one shelf) vary in shape (rectangular or circular), surface appearance (wood grain or metal), and size.
Four table variants are used for training; the remaining two tables and the shelf are held out for out-of-distribution evaluation. We vary the initial radial distance and approach angle, the target appearance, and the scene layout. Each trajectory receives instruction paraphrases that preserve the target object and action, such as ``Walk ahead, stop at the table.''  To scale data acquisition, we also combine robot-collected and human-collected trajectories, with human operators matching the robot's camera height and motion dynamics.

We adopt HOMIE~\cite{ben2024homie}, trained with PPO~\cite{schulman2017proximal} and an upper-body pose curriculum for balance robustness, as the shared WBC across navigation, calibration, and manipulation.
It generates motor actions from proprioceptive state history, high-level commands, and the previous motor action.

\subsection{Posture Calibration for Pre-Manipulation}
Navigation alone does not guarantee a manipulation-ready posture for dexterous, contact-rich tasks.
We therefore define a manipulation-ready posture as one where the target object is centered within a 3D workspace region where the arm end-effector (EE) achieves low inverse-kinematics (IK) error under the utilized IK solver. 

To realize a manipulation-ready posture, we introduce a lightweight closed-loop posture calibration module that refines the robot base pose using online object pose feedback. Let $O$ be the nonempty set of task-relevant objects and $|O|$ its cardinality. At each time step, we extract their robot-frame positions $\{(x_o,y_o,z_o)\}_{o\in O}$, with forward, lateral, and vertical coordinates $x_o,y_o,z_o$. We select the object with the largest absolute forward offset and use its coordinate:
$o^\star=\arg\max_{o\in O}|x_o|$, $x=x_{o^\star}$.
The lateral and vertical coordinates are averaged across objects: $(y,z)=\tfrac{1}{|O|}\sum_{o\in O}(y_o,z_o)$.

\begin{figure}[!t]
    \centering
    \includegraphics[width=\columnwidth]{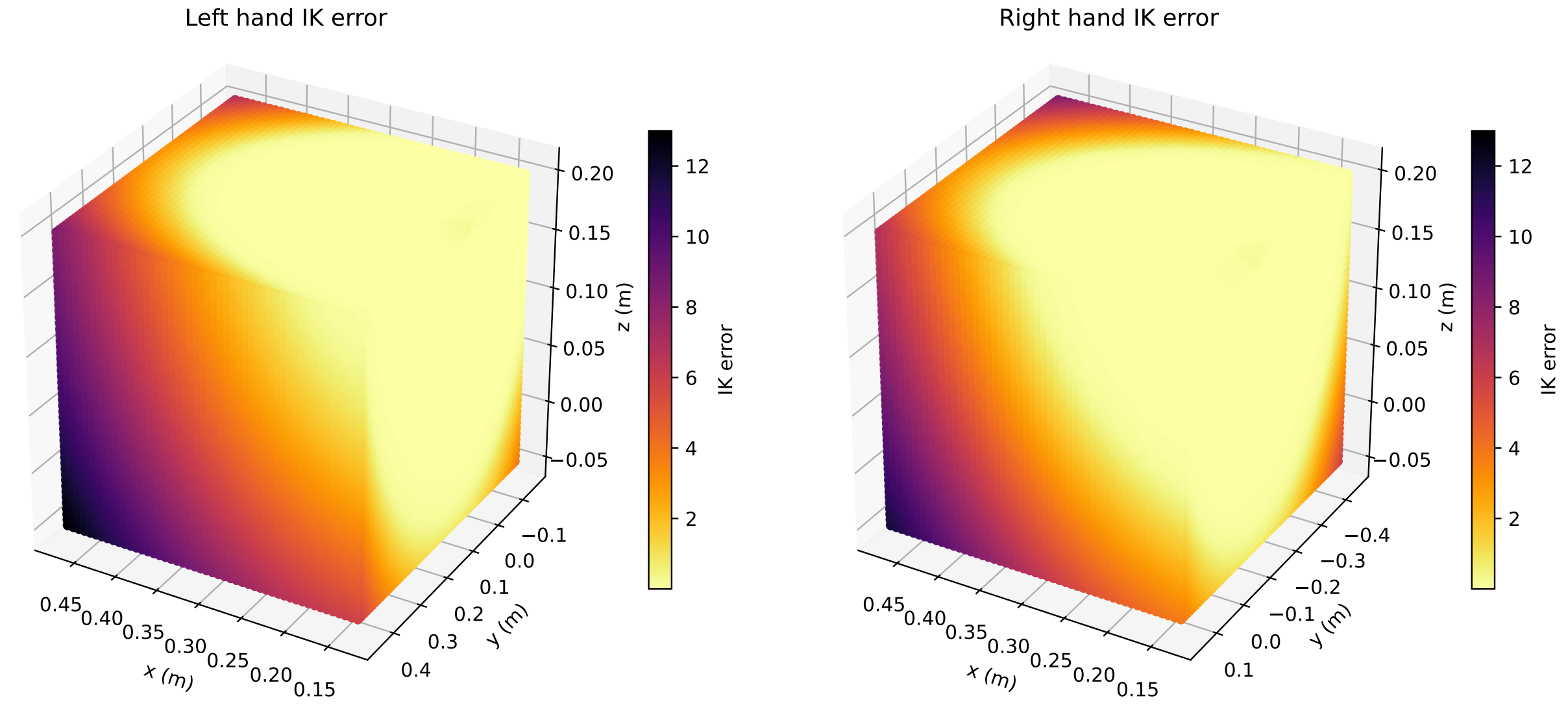}
    \caption{End-effector IK error for the left and right arms with a fixed torso.}
    \label{fig:ik-error-thick}
\end{figure}

Our module first adjusts the robot’s height to eliminate the vertical deviation $(z_{\mathrm{ref}}-z)$, then performs closed-loop regulation by driving the object displacement $(x,y)$ toward a reference $(x_{\mathrm{ref}}, y_{\mathrm{ref}})$, generating planar velocity commands $(v_x, v_y)$ as bounded functions of the displacement errors and their temporal accumulation. 
$z_{\mathrm{ref}}$ is set to the average height of the target objects in the egocentric videos. Fig.~\ref{fig:ik-error-thick} shows the end-effector IK error distributions with a fixed torso. We choose $(x_{\mathrm{ref}},y_{\mathrm{ref}})=(0.4,0)$\,m: $0.4$\,m is the farthest forward position with tolerant IK error, and $0$ centers the low-error lateral region.
To suppress unnecessary corrections near the target, we use position-error deadband thresholds $\varepsilon_x,\varepsilon_y>0$: $|x-x_{\mathrm{ref}}|\le\varepsilon_x$, $|y-y_{\mathrm{ref}}|\le\varepsilon_y$. We limit planar velocity commands by $v_i\leftarrow\mathrm{clip}(v_i,-v_i^{\max},v_i^{\max})$, where $i\in\{x,y\}$, $v_i^{\max}>0$ is the corresponding speed limit. The resulting commands are issued at a fixed rate and continuously updated until the robot reaches a manipulation-ready workspace.

\subsection{Wrist and Dexterous Hand Motion Extraction}

After the above two stages, the problem reduces to generating precise arm and hand motions to accomplish the task. We extract wrist trajectories and finger motions from one-shot RGB-D egocentric demonstrations recorded with the G1 humanoid's head-mounted D435i. Demonstrations are collected within the G1 arm-hand workspace; when the torso restricts smooth movement of a single demonstrator, two operators can demonstrate the left- and right-hand motions collaboratively. We convert these priors into verified keyframes of EE poses and hand motor commands.

\textbf{Human Wrist Motion Extraction.}
For a $T$-frame single-view RGB-D sequence $\{I_t\}_{t=1}^{T}$, a MANO-based reconstruction model~\cite{Potamias2025WiLoR} estimates left/right hand meshes ($\diamond\in\{L,R\}$); here $t$ indexes video frames. MANO's global axis-angle orientation~\cite{romero2017embodied} yields a camera-frame wrist rotation $h^\diamond_{r,t}\in\mathrm{SO}(3)$. Back-projecting the wrist pixel $(u,v)$ with depth $d$ and camera intrinsics gives the camera-frame position $h^\diamond_{p,t}\in\mathbb{R}^3$. Eye-to-hand calibration provides camera-to-robot rotation $R_{\mathrm{cam}}^{\mathrm{robot}}\in\mathrm{SO}(3)$ and translation $\mathbf{t}_{\mathrm{cam}}^{\mathrm{robot}}\in\mathbb{R}^3$. The robot-frame EE poses are
\[
a^\diamond_{r,t}=R_{\mathrm{cam}}^{\mathrm{robot}}h^\diamond_{r,t},
\qquad
a^\diamond_{p,t}=R_{\mathrm{cam}}^{\mathrm{robot}}h^\diamond_{p,t}
+\mathbf{t}_{\mathrm{cam}}^{\mathrm{robot}}.
\]
Subscripts $r$ and $p$ denote rotation and position, respectively.

\textbf{Keyframe-Based Motion Refinement.}
Following prior work~\cite{james2022coarse,zhou2025you}, we compress noisy, redundant trajectories into $K\ll T$ salient keyframes indexed by $k$, $\{a^{\diamond}_{r,k},a^{\diamond}_{p,k}\}_{k=1}^K$. 
Keyframes are selected at moments of significant motion change, typically when 
(i) a dexterous hand is interacting with the object, or
(ii) the EE twist
\[
\begin{bmatrix}
\mathbf{v}^{\diamond}_t\\
\boldsymbol{\omega}^{\diamond}_t
\end{bmatrix}
\approx \frac{1}{\Delta t}
\left[\log\left((G^{\diamond}_{t-1})^{-1}
G^{\diamond}_t\right)\right]^{\vee}
\]
attains a local extremum.
Here $G_t^\diamond=
\left[\begin{smallmatrix}
a^\diamond_{r,t} & a^\diamond_{p,t}\\
\mathbf{0}^{\top} & 1
\end{smallmatrix}\right]\in\mathrm{SE}(3)$
is the homogeneous EE pose, and $\Delta t$ is the interval
between consecutive video frames.
The linear and angular velocities
$\mathbf{v}_t^\diamond,\boldsymbol{\omega}_t^\diamond
\in\mathbb{R}^3$ are expressed in the EE frame at $t-1$. The Lie group logarithm $\log:\mathrm{SE}(3)\to\mathfrak{se}(3)$ and vee operator $(\cdot)^\vee:\mathfrak{se}(3)\to\mathbb{R}^6$ convert the relative transformation into its linear and angular twist components.

\textbf{Dexterous Retargeting.}
At each keyframe $k$, we geometrically infer $11$ joint angles in degrees from the $21$ hand landmarks detected by MediaPipe~\cite{zhang2020mediapipe}. These cover the proximal and distal joints of all five fingers and the thumb metacarpal joint. Following the Revo2 URDF, we retain $6$ independently controlled joints; the other $5$ distal joints follow predefined mimic relations with their proximal joints. The motor command for each hand is
\[
a^\diamond_{m,k}=\left[\left\lfloor 1000\,\mathrm{clip}\!\left(\frac{h^\diamond_{m,j,k}}{h^{\max}_{m,j}},0,1\right)\right\rfloor\right]_{j\in\mathcal{J}_6},
\]
where $m$ denotes hand motion, $\mathcal{J}_6$ is the ordered set of controlled joint indices, $h^\diamond_{m,j,k}$ is the angle of joint $j$ on side $\diamond$, and $h^{\max}_{m,j}$ is its maximum angle. The floor operator $\lfloor\cdot\rfloor$ yields integer commands in $\{0,\ldots,1000\}$. The sequence $\{a^\diamond_{m,k}\}_{k=1}^{K}$ drives dexterous hand control.

\begin{figure}[t]
    \centering
    \includegraphics[width=\linewidth,page=3]{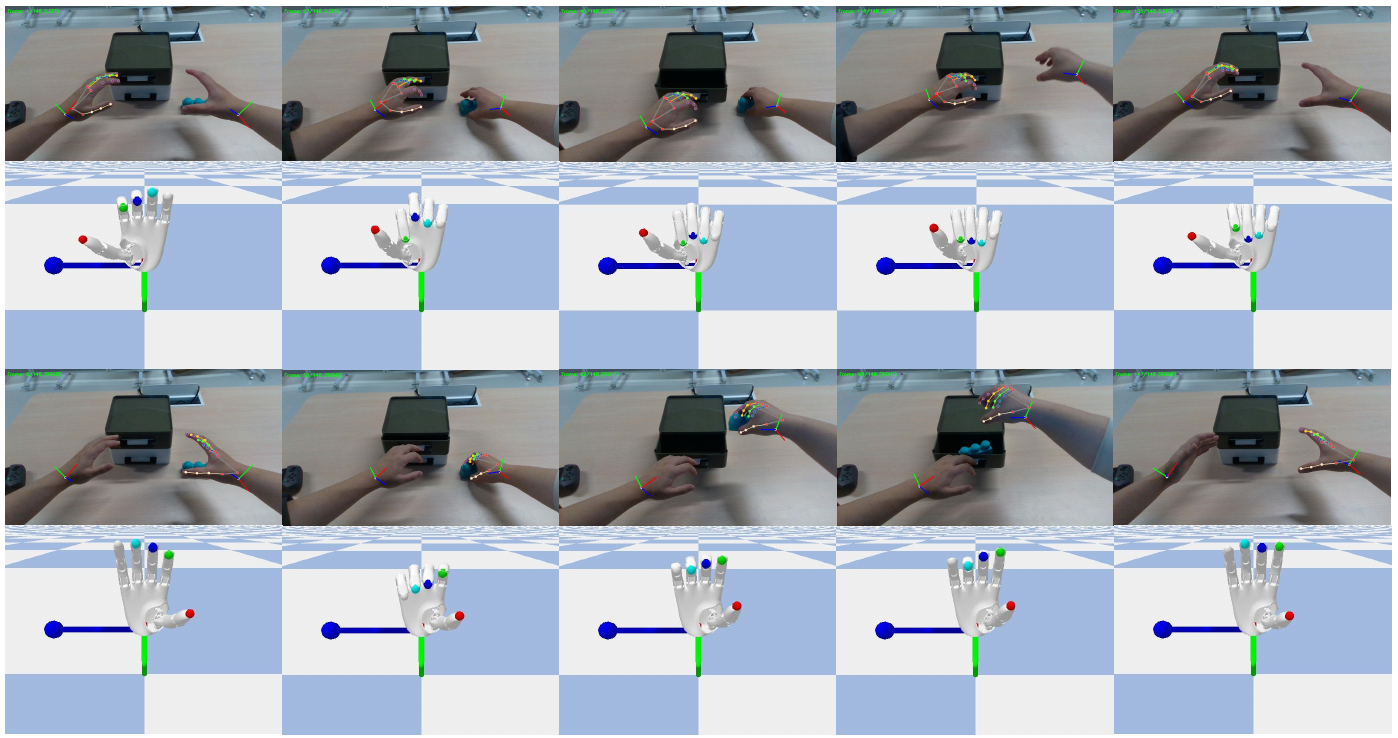}
    \caption{Visualization of wrist pose detection and dexterous retargeting for the \texttt{Pull Drawer} task.}
    \label{fig:pull-drawer-qualitative}
\end{figure}

\textbf{Robot-Space Verification.} Robot-frame trajectories may contain unreachable targets or self-collisions. We validate keyframes offline using IK and hand-level control constraints, retaining reachable, collision-free actions. The verified per-arm sequences are $A^\diamond=\{(a^{\diamond}_{r,k},a^{\diamond}_{p,k},a^{\diamond}_{m,k})\}_{k=1}^{K}$ for $\diamond\in\{L,R\}$, where $K$ is the number of verified actions. Fig.~\ref{fig:pull-drawer-qualitative} illustrates an example action sequence for the \texttt{Pull Drawer} task.

\begin{table*}[!t]
\centering
\renewcommand{\arraystretch}{1.0}
\setlength{\tabcolsep}{1pt}
\setlength{\aboverulesep}{0.35ex}
\setlength{\belowrulesep}{0.35ex}

\caption{Step-wise success rates (successes/trials) and average number of completed steps (Avg. Len.) across navigation (Nav.), posture
calibration (Calib.), and manipulation (Manip.).
M.: translation-only approach; M.+T.: turning required.
w/o Adapt.: original VLN policy.
Colors denote the left (\textcolor{RoyalBlue}{blue}),
right (\textcolor{Orange!80!black}{orange}),
and both arms (\textcolor{purple!68}{purple}).}
\label{tab:long-horizon}

\footnotesize

\begin{tabular*}{\linewidth}
{@{\extracolsep{\fill}}llcccccccc@{}}
\toprule
\multicolumn{2}{l}{\texttt{Hand Over}} &
Stage 1 &
Stage 2 &
\multicolumn{5}{c}{Stage 3: Manip.} &
\multirow{2}{*}{\textbf{Avg. Len.}} \\
\cmidrule{3-3}\cmidrule{4-4}\cmidrule{5-9}
Methods & Distance & Nav. & Calib. &
\textcolor{Orange!80!black}{pick cup} &
\textcolor{Orange!80!black}{close to left hand} &
\textcolor{purple!68}{hand over} &
\textcolor{RoyalBlue}{move to tray} &
\textcolor{RoyalBlue}{place cup} & \\
\midrule

\Method{} & $3$\,m (M.)
& 10/10 & 9/10
& 9/10 & 9/10 & 9/10 & 9/10 & 9/10
& \cellcolor{gray!20}6.4 \\

\Method{} w/o Adapt. & $3$\,m (M.)
& 1/10 & 1/10
& 1/10 & 0/10 & 0/10 & 0/10 & 0/10
& \cellcolor{gray!20}0.3 \\

\Method{} & $5$\,m (M.)
& 7/10 & 7/10
& 7/10 & 7/10 & 6/10 & 6/10 & 6/10
& \cellcolor{gray!20}4.6 \\
\midrule

\Method{} & $3$\,m (M.+T.)
& 7/10 & 6/10
& 5/10 & 6/10 & 5/10 & 5/10 & 5/10
& \cellcolor{gray!20}3.9 \\

\Method{} w/o Adapt. & $3$\,m (M.+T.)
& 0/10 & 0/10
& 0/10 & 0/10 & 0/10 & 0/10 & 0/10
& \cellcolor{gray!20}0.0 \\

\Method{} & $5$\,m (M.+T.)
& 4/10 & 4/10
& 4/10 & 3/10 & 3/10 & 3/10 & 3/10
& \cellcolor{gray!20}2.4 \\

\bottomrule
\end{tabular*}

\par\vspace{2pt}

\begin{tabular*}{\linewidth}
{@{\extracolsep{\fill}}lccccccccc@{}}
\texttt{Pour Water} &
Stage 1 &
Stage 2 &
\multicolumn{6}{c}{Stage 3: Manip.} &
\multirow{2}{*}{\textbf{Avg. Len.}} \\
\cmidrule{2-2}\cmidrule{3-3}\cmidrule{4-9}
Distance & Nav. & Calib. &
\textcolor{RoyalBlue}{pick cup} &
\textcolor{Orange!80!black}{pick bottle} &
\textcolor{purple!68}{close to each other} &
\textcolor{Orange!80!black}{pour water} &
\textcolor{Orange!80!black}{place bottle} &
\textcolor{RoyalBlue}{place cup} & \\
\midrule

$3$\,m (M.)
& 10/10 & 10/10
& 10/10 & 10/10 & 10/10 & 10/10 & 10/10 & 10/10
& \cellcolor{gray!20}8.0 \\

$5$\,m (M.)
& 8/10 & 7/10
& 7/10 & 7/10 & 7/10 & 7/10 & 7/10 & 7/10
& \cellcolor{gray!20}5.7 \\
\midrule

$3$\,m (M.+T.)
& 6/10 & 6/10
& 6/10 & 6/10 & 6/10 & 6/10 & 6/10 & 6/10
& \cellcolor{gray!20}4.8 \\

$5$\,m (M.+T.)
& 5/10 & 5/10
& 5/10 & 5/10 & 5/10 & 4/10 & 4/10 & 4/10
& \cellcolor{gray!20}3.7 \\

\bottomrule
\end{tabular*}

\end{table*}

\subsection{One-shot Policy Generalization from Egocentric Video}\label{main:exp-gen}

A single demonstration specifies task-critical EE approach geometry and hand-object contact patterns. \Method{} preserves these priors while adapting approach, alignment, and corrective motion through visual and tactile feedback, 
enabling generalization to new object poses and scene layouts, and recovery from execution disturbances.

\textbf{Generalization across Object Poses.}
We condition actions on online object poses estimated by FoundationPose++~\cite{Wenhao_Yan_and_Jie_Chu_FoundationPose_2025}, which extends FoundationPose~\cite{wen2024foundationpose} with 2D tracking and Kalman filtering for temporally consistent multi-object 6D pose estimation. 
We use LISA ~\cite{lai2023lisa} to obtain the object masks required for pose estimation.
At interaction keyframe $k$, let $T^{\diamond}_{o_i,k}\in\mathrm{SE}(3)$ and $T^{\diamond}_{a,k}=(a^{\diamond}_{r,k},a^{\diamond}_{p,k})\in\mathrm{SE}(3)$ denote the demonstrated poses of object $o_i$ and the EE of hand $\diamond$, respectively, in the robot frame. Their relative transform
$\Delta T^{\diamond}_{k}=(T^{\diamond}_{o_i,k})^{-1}T^{\diamond}_{a,k}$
expresses the EE pose in the object frame. Given a newly estimated pose $\hat{T}^{\diamond}_{o_i,k}$ of the same object in a novel rollout, the adapted EE pose is
$\hat{T}^{\diamond}_{a,k}=\hat{T}^{\diamond}_{o_i,k}\,\Delta T^{\diamond}_{k}=(\hat{a}^{\diamond}_{r,k},\hat{a}^{\diamond}_{p,k})$.
Updating the corresponding EE poses in $A^\diamond$ yields $\hat{A}^\diamond$, preserving the demonstrated object-relative interaction geometry across object pose variations.
Objects of similar shape and scale can replace the demonstrated object, enabling cross-object generalization. We also support spatial edits for multiple affordances, such as selecting different drawer handles by adjusting $z$ in $\hat{a}^{\diamond}_{p,k}$.

\textbf{Generalization across Scene Variations.} The G1's left-right symmetry allows the demonstrated EE trajectory $A^\diamond$ to be reflected across the sagittal plane, exchanging the active hand under symmetric task layouts without another demonstration. FoundationPose++~\cite{Wenhao_Yan_and_Jie_Chu_FoundationPose_2025} supports online re-grounding under changes in lighting, scene layout, and surface appearance.

\textbf{Tactile Grasp Regulation.} 
Tactile feedback regulates contact forces following the demonstrated feedforward hand closure.
Each finger $i\in\{1,\ldots,5\}$ estimates a fingertip contact force $\mathbf{f}_i=[f_{n,i},\mathbf{f}_{t,i}]$, where $f_{n,i}\ge0$ is the normal force magnitude and $\mathbf{f}_{t,i}\in\mathbb{R}^2$ is the tangential force vector. A binary indicator $c_i=\mathbb{I}(f_{n,i}>\tau_n)$ detects contact using a predefined threshold $\tau_n$, where $\mathbb{I}$ equals $1$ when its condition holds and $0$ otherwise. We adopt a heuristic stability criterion requiring at least $M$ active contacts ($\sum_i c_i\ge M$). 
Once this criterion is satisfied, we estimate the contact stiffness
as $\kappa = \operatorname{mean}(\Delta f_{n,i}/\Delta x_{n,i})$,
averaging the ratio of normal-force change $\Delta f_{n,i}$
to normal-indentation change $\Delta x_{n,i}$ over contacting fingers.
The desired normal grasp force is then computed as $f_n^\star=\mathrm{clip}(\alpha\kappa+\beta,f_{\min},f_{\max})$, where $\alpha,\beta$ are tuned gains and $[f_{\min},f_{\max}]$ are bounding limits. To prevent slipping under a Coulomb friction model, each finger's target normal force satisfies $f_{n,i}^\star\ge \lVert\mathbf{f}_{t,i}\rVert/(\mu_i-\delta)$, where $\mu_i$ is the local friction coefficient and $\delta>0$ is a safety margin. 

\begin{table*}[!t]
\renewcommand{\arraystretch}{1.08}
\setlength{\tabcolsep}{1pt}
\centering
\caption{Quantitative performance comparisons on five long-horizon whole-body manipulation tasks. We report step-wise success rates and the average number of completed steps (Avg. Len.). Bold entries denote the highest Avg. Len. The two press--release cycles in \texttt{Manipulate Pipette} correspond to aspiration and dispensing.}
\label{tab:object-pose}

\footnotesize
\begin{tabular*}{\linewidth}{@{\extracolsep{\fill}} l cccccc ccccccc @{}}
    \toprule
    \multirow{2}{*}{Methods}
    & \multicolumn{6}{c}{\texttt{Hand Over}}
    & \multicolumn{7}{c}{\texttt{Pour Water}} \\
    \cmidrule(lr){2-7}
    \cmidrule(lr){8-14}
    & \textcolor{Orange!80!black}{\makecell{pick\\cup}}
    & \textcolor{Orange!80!black}{\makecell{close to\\left hand}}
    & \textcolor{purple!68}{\makecell{hand\\over}}
    & \textcolor{RoyalBlue}{\makecell{move to\\tray}}
    & \textcolor{RoyalBlue}{\makecell{place\\cup}}
    & \textbf{\makecell{Avg.\\Len.}}
    & \textcolor{RoyalBlue}{\makecell{pick\\cup}}
    & \textcolor{Orange!80!black}{\makecell{pick\\bottle}}
    & \textcolor{purple!68}{\makecell{close to\\each other}}
    & \textcolor{Orange!80!black}{\makecell{pour\\water}}
    & \textcolor{Orange!80!black}{\makecell{place\\bottle}}
    & \textcolor{RoyalBlue}{\makecell{place\\cup}}
    & \textbf{\makecell{Avg.\\Len.}} \\
    \midrule

    $\pi_{0.5}$
    & 29/33 & 26/33 & 18/33 & 16/33 & 15/33
    & \cellcolor{gray!20}3.15
    & 26/33 & 28/33 & 25/33 & 20/33 & 18/33 & 17/33
    & \cellcolor{gray!20}4.06 \\

    YOTO
& 30/33 & 29/33 & 27/33 & 27/33 & 27/33
& \cellcolor{gray!20}4.24
& 32/33 & 30/33 & 30/33 & 29/33 & 29/33 & 29/33
& \cellcolor{gray!20}5.42 \\
    OKAMI    
& 25/33 & 25/33 & 17/33 & 15/33 & 15/33    
& \cellcolor{gray!20}2.94    
& 24/33 & 26/33 & 24/33 & 20/33 & 18/33 & 18/33    
& \cellcolor{gray!20}3.94 \\
    \Method{} w/o Tact.    
& 28/33 & 28/33 & 26/33 & 26/33 & 26/33    
& \cellcolor{gray!20}4.06    
& 29/33 & 31/33 & 28/33 & 27/33 & 27/33 & 27/33    
& \cellcolor{gray!20}5.12 \\
    \Method{} (Ours)    
& 31/33 & 31/33 & 31/33 & 30/33 & 30/33    
& \cellcolor{gray!20}\textbf{4.64}    
& 32/33 & 31/33 & 30/33 & 30/33 & 30/33 & 30/33    
& \cellcolor{gray!20}\textbf{5.55} \\

    \bottomrule
\end{tabular*}

\vspace{2pt}

\begin{tabular*}{\linewidth}{@{\extracolsep{\fill}} l ccccc cccccc cccccc @{}}
    \multirow{2}{*}{Methods}
    & \multicolumn{5}{c}{\texttt{Pull Drawer}}
    & \multicolumn{6}{c}{\texttt{Open Lid}}
    & \multicolumn{6}{c}{\texttt{Manipulate Pipette}} \\
    \cmidrule(lr){2-6}
    \cmidrule(lr){7-12}
    \cmidrule(lr){13-18}
    & \textcolor{RoyalBlue}{\makecell{pull\\drawer}}
    & \textcolor{Orange!80!black}{\makecell{pick\\block}}
    & \textcolor{Orange!80!black}{\makecell{place\\block}}
    & \textcolor{RoyalBlue}{\makecell{push\\drawer}}
    & \textbf{\makecell{Avg.\\Len.}}
    & \textcolor{Orange!80!black}{\makecell{lift\\lid}}
    & \textcolor{RoyalBlue}{\makecell{pick\\toy}}
    & \textcolor{Orange!80!black}{\makecell{move lid\\away}}
    & \textcolor{RoyalBlue}{\makecell{place toy\\in box}}
    & \textcolor{Orange!80!black}{\makecell{place\\lid}}
    & \textbf{\makecell{Avg.\\Len.}}
    & \textcolor{Orange!80!black}{\makecell{pick\\pipette}}
    & \textcolor{Orange!80!black}{\makecell{press\\button}}
    & \textcolor{Orange!80!black}{\makecell{release\\button}}
    & \textcolor{Orange!80!black}{\makecell{press\\button}}
    & \textcolor{Orange!80!black}{\makecell{release\\button}}
    & \textbf{\makecell{Avg.\\Len.}} \\
    \midrule

    $\pi_{0.5}$
    & 16/33 & 28/33 & 13/33 & 11/33
    & \cellcolor{gray!20}2.06
    & 11/33 & 29/33 & 9/33 & 8/33 & 8/33
    & \cellcolor{gray!20}1.97
    & 18/33 & 5/33 & 5/33 & 0/33 & 0/33
    & \cellcolor{gray!20}0.85 \\

    YOTO
& 26/33 & 32/33 & 23/33 & 22/33
& \cellcolor{gray!20}3.12
& 21/33 & 30/33 & 17/33 & 17/33 & 17/33
& \cellcolor{gray!20}3.09
& 26/33 & 16/33 & 16/33 & 15/33 & 15/33
& \cellcolor{gray!20}2.67 \\
    OKAMI    
& 20/33 & 26/33 & 18/33 & 17/33    
& \cellcolor{gray!20}2.45    
& 13/33 & 25/33 & 11/33 & 11/33 & 10/33    
& \cellcolor{gray!20}2.12    
& 15/33 & 8/33 & 7/33 & 2/33 & 2/33    
& \cellcolor{gray!20}1.03 \\
    \Method{} w/o Tact.    
& 24/33 & 29/33 & 22/33 & 22/33    
& \cellcolor{gray!20}2.94    
& 19/33 & 29/33 & 17/33 & 16/33 & 16/33    
& \cellcolor{gray!20}2.94    
& 24/33 & 17/33 & 17/33 & 14/33 & 14/33    
& \cellcolor{gray!20}2.61 \\
    \Method{} (Ours)    
& 29/33 & 31/33 & 26/33 & 26/33    
& \cellcolor{gray!20}\textbf{3.39}    
& 25/33 & 31/33 & 22/33 & 21/33 & 20/33    
& \cellcolor{gray!20}\textbf{3.61}    
& 28/33 & 23/33 & 23/33 & 21/33 & 21/33    
& \cellcolor{gray!20}\textbf{3.52} \\

    \bottomrule
\end{tabular*}
\end{table*}

\textbf{Generalization to Failure Cases.}
To verify interaction outcomes and guide recovery, Praxis
uses Qwen3.8-27B~\cite{qwen38_27b} as an online verifier at each manipulation
keyframe $k$:
\[
\psi_{t,k}
= C_\zeta(\Phi_t,\mathcal{D}_k)
\in \{1,0,\mathrm{uncertain}\},
\]
where $\Phi_t$ denotes the robot's current egocentric
observations and $\mathcal{D}_k$ the corresponding reference
from the human demonstration. 
A successful outcome ($1$) advances execution to the next keyframe, whereas a failed outcome ($0$) triggers updated object perception and corrective arm–hand replanning. An uncertain outcome triggers additional observations and reassessment before execution advances.
The same one-shot demonstration thus supports
both motion transfer and online execution supervision.

\begin{figure*}[!t]
    \centering
    \includegraphics[width=\linewidth,pagebox=cropbox]{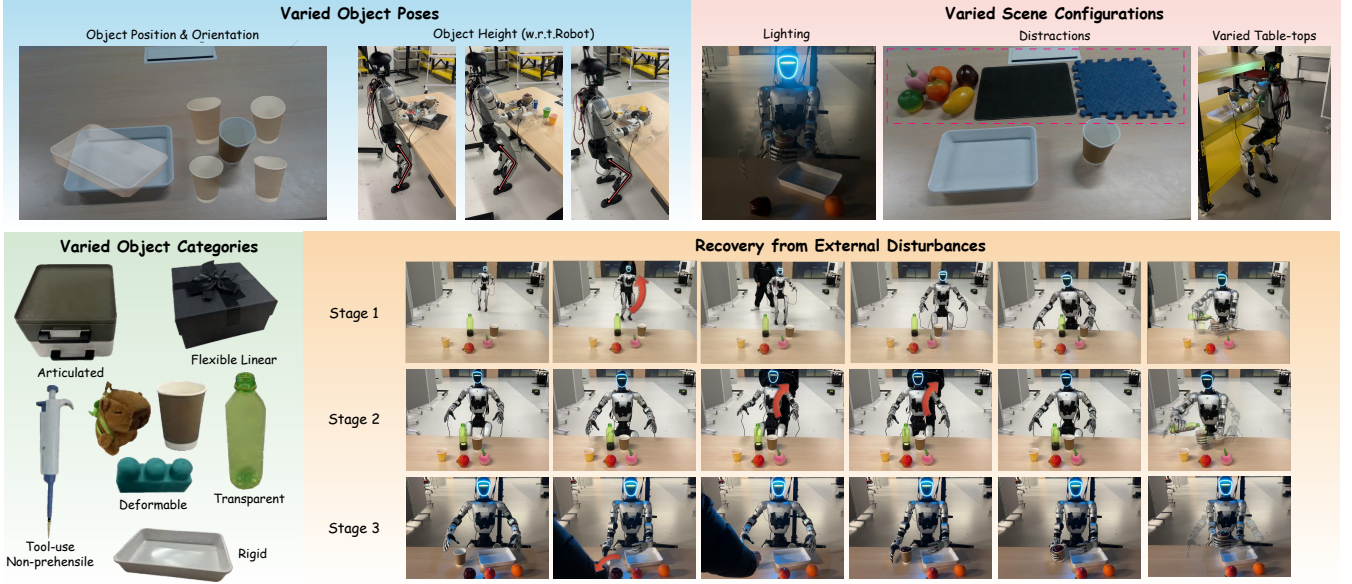}
    \vspace{-0.2in}
    \caption{Visualization of varied object poses, scene configurations, and object categories, and failure recovery across three stages.}
    \label{fig:generalizability}
\end{figure*}

\section{Experiments}\label{main:exp}

In the experiments, we seek to address the following research questions.
\begin{itemize}
    \item \textit{Q1}: How effective is the proposed system in whole-body manipulation tasks?
    \item \textit{Q2}: How well does the system generalize across variations in object poses and scene configurations, and how does tactile feedback contribute to contact-sensitive manipulation?
    \item \textit{Q3}: How effectively does the system recover from external interference during navigation, posture calibration, and manipulation?
\end{itemize}

\subsection{Experimental Setup}

\subsubsection{Tasks}

We evaluate \Method{} across its three operational phases: navigation, posture calibration, and manipulation. To assess the system's robustness during navigation and calibration, we evaluate initial distances of 3 m and 5 m from the workspace. We benchmark the performance at each distance under two distinct locomotion protocols: a direct linear approach versus a sequential `orient-then-advance' strategy toward the workspace.
We evaluate the manipulation phase on five real-world whole-body manipulation tasks:
Hand Over (paper cup and tray), Pour Water (bottle and cup), Pull Drawer (drawer and block), Open Lid (lidded box and toy), and Manipulate Pipette (pipette, rack, and tray). They contain $4$–$6$ manipulation steps, listed in Table \ref{tab:object-pose}.
The objects manipulated in these tasks span rigid, transparent, articulated, deformable, flexible strip-like, and non-prehensile categories. Solving them requires a diverse set of primitive skills, including pick-and-place, re-orient, pull-and-push, lift, and dexterous finger motions such as open, clench, flex, pinch, press, and release. Moreover, all tasks require long-horizon dexterous manipulation, involving multiple sequential or parallel steps.

\subsubsection{Baselines}

We consider three complementary baselines: $\pi_{0.5}$~\cite{intelligence2025pi}, a vision-language-action model; YOTO, a one-shot bimanual manipulation framework; and OKAMI~\cite{li2024okami}, a single-video imitation method. YOTO extracts bimanual keyframes from a single human demonstration. Originally developed for parallel-jaw grippers, it uses the mean fingertip position as the EE position reference and represents grasping with a binary open/closed state. Consequently, this gripper-oriented representation captures EE motion and binary grasp states but does not preserve the fine-grained hand motion priors encoded in the demonstration. To better isolate the effect of this representational difference, we also equip the YOTO baseline with the same tactile sensing as Praxis. OKAMI extracts a manipulation plan from a single third-person human demonstration and performs object-aware retargeting of body motions and hand poses to accommodate new object locations. Both YOTO and OKAMI use the same object pose estimation method as \Method{}.

The $\pi_{0.5}$ baseline generates upper-body actions while sharing the same lower-body controller as \Method{}. We collect $150$ teleoperated trajectories per task using Meta Quest 3 and finetune $\pi_{0.5}$ for $5000$ steps with batch size $128$. The action/state vector has $26$ dimensions: $7$ per arm and $6$ per hand. Images and states are aligned and stored in LeRobot~\cite{cadene2024lerobot} v3.0 format.

\subsubsection{Metrics}

For the manipulation comparisons in Table~\ref{tab:object-pose}, we evaluate $\pi_{0.5}$, YOTO, OKAMI, \Method{} w/o Tact. (without tactile feedback), and \Method{} with $33$ trials per task. 
For a more granular comparison, we report the mean number of successfully completed substeps per trial (Avg. Len.; following CALVIN~\cite{mees2022calvin}),
and step-wise success rates (the proportion of trials in which each step is successfully completed).
Navigation and calibration are counted separately in Table \ref{tab:long-horizon} and jointly in Tables \ref{tab:recovery}–\ref{tab:vln-calib-ablation}.
Aggregate success rates are computed over the specified tasks and evaluation conditions.
We exclude trials affected by hardware anomalies, such as G1 or Revo2 failures due to low battery or overheating.
For manipulation tasks, we use the intersection of the low-IK-error workspace in Fig.~\ref{fig:ik-error-thick} and the G1 head-camera FOV to define the feasible object placement region. 
To preserve sufficient manipulation clearance, we also avoid placing objects too close to the dexterous hand, especially near the back of the hand.

\begin{table}[!t]
\centering
\renewcommand{\arraystretch}{1.0}
\setlength{\tabcolsep}{3pt}
\setlength{\aboverulesep}{0.35ex}
\setlength{\belowrulesep}{0.35ex}

\caption{Step-wise success rates and average number of completed steps (Avg. Len.) on recovery cases across all stages for \texttt{Pour Water}.}
\label{tab:recovery}

\footnotesize
\begin{tabular*}{\columnwidth}
{@{\extracolsep{\fill}}lccc@{}}
\toprule
\multirow{2}{*}{Step} &
\multicolumn{3}{c}{Interference at} \\
\cmidrule{2-4}
& Stage 1\&2 & Stage 2 & Stage 3 \\
\midrule
Stage 1\&2 & 8/10 & 9/10 & 10/10 \\
\midrule
\multicolumn{4}{l}{Stage 3: Manip.} \\

\textcolor{RoyalBlue}{pick cup}
& 7/10 & 9/10 & 10/10 \\
\textcolor{Orange!80!black}{pick bottle}
& 8/10 & 9/10 & 10/10 \\
\textcolor{purple!68}{close to each other}
& 7/10 & 9/10 & 10/10 \\
\textcolor{Orange!80!black}{pour water}
& 7/10 & 9/10 & 9/10 \\
\textcolor{Orange!80!black}{place bottle}
& 7/10 & 9/10 & 9/10 \\
\textcolor{RoyalBlue}{place cup}
& 7/10 & 9/10 & 9/10 \\

\midrule
\textbf{Avg. Len.}
& \cellcolor{gray!20}5.1
& \cellcolor{gray!20}6.3
& \cellcolor{gray!20}6.7 \\
\bottomrule
\end{tabular*}
\end{table}

\begin{table}[!t]
\centering
\renewcommand{\arraystretch}{1.0}
\setlength{\tabcolsep}{3pt}
\setlength{\aboverulesep}{0.35ex}
\setlength{\belowrulesep}{0.35ex}

\caption{Step-wise success rates and average number of completed steps (Avg. Len.) under lateral interference across all three stages
for \texttt{Pour Water}.
Stage 1 Avg. Step is the average number of inference steps
during navigation. Calib. denotes posture calibration.}
\label{tab:vln-calib-ablation}

\footnotesize
\begin{tabular*}{\columnwidth}
{@{\extracolsep{\fill}}lcc@{}}
\toprule
\multirow{2}{*}{Step} &
\multicolumn{2}{c}{Method} \\
\cmidrule{2-3}
& \Method{} w/o Calib. & \Method{} (Ours) \\
\midrule
Stage 1\&2 & 2/10 & 8/10 \\
\midrule
\multicolumn{3}{l}{Stage 3: Manip.} \\

\textcolor{RoyalBlue}{pick cup}
& 1/10 & 8/10 \\
\textcolor{Orange!80!black}{pick bottle}
& 2/10 & 8/10 \\
\textcolor{purple!68}{close to each other}
& 1/10 & 8/10 \\
\textcolor{Orange!80!black}{pour water}
& 1/10 & 8/10 \\
\textcolor{Orange!80!black}{place bottle}
& 1/10 & 8/10 \\
\textcolor{RoyalBlue}{place cup}
& 1/10 & 7/10 \\

\midrule
\textbf{Avg. Len.}
& \cellcolor{gray!20}0.9
& \cellcolor{gray!20}\textbf{5.5} \\
\midrule
Stage 1 Avg. Step & 13.2 & 13.5 \\
\bottomrule
\end{tabular*}
\end{table}

\subsection{Experimental Results}


\textbf{\textit{(Q1)} Whole-body manipulation effectiveness.}
Table~\ref{tab:long-horizon} reports long-distance, long-horizon performance. At $3$\,m without turning, \Method{} achieves perfect navigation success and high manipulation reliability in \texttt{Hand Over}, and $100\%$ success across all stages in \texttt{Pour Water}. Using the original VLN model without manipulation-region adaptation almost eliminates \texttt{Hand Over} completion. In the challenging $5$\,m turning setting, the two tasks achieve $30\%$ and $40\%$ success with average lengths $2.4$ and $3.7$. Averaging across tasks and approach protocols gives $75\%$ success at $3$\,m and $50\%$ at $5$\,m. The adapted VLN policy and object-centric calibration support the transition to manipulation-ready postures.

\begin{table}[!tb]
\centering
\caption{Success rates (\%) across five tasks, with $33$ trials per task. Mean denotes the average success rate across the five tasks.}
\label{tab:success-summary}
\footnotesize
\setlength{\tabcolsep}{2pt}
\renewcommand{\arraystretch}{1.12}
\begin{tabular*}{\linewidth}{@{\extracolsep{\fill}}lccccc@{}}
\toprule
Task & $\pi_{0.5}$ & YOTO & OKAMI & \makecell{\Method{}\\w/o Tact.} & \makecell{\Method{}\\(Ours)} \\
\midrule
Hand Over & 45.45 & 81.82 & 45.45 & 78.79 & \textbf{90.91} \\
Pour Water & 51.52 & 87.88 & 54.55 & 81.82 & \textbf{90.91} \\
Pull Drawer & 33.33 & 66.67 & 51.52 & 66.67 & \textbf{78.79} \\
Open Lid & 24.24 & 51.52 & 30.30 & 48.48 & \textbf{60.61} \\
Manipulate Pipette & 0.00 & 45.45 & 6.06 & 42.42 & \textbf{63.64} \\
\midrule
Mean & 30.91 & 66.67 & 37.58 & 63.64 & \textbf{76.97} \\
\bottomrule
\end{tabular*}
\end{table}
\textbf{\textit{(Q2)} Generalization of \Method{} under diverse environmental variations.} 
To validate our method as a generalizable policy, we introduce systematic variations in the initial object pose for each task. Specifically, we randomize the planar position, height relative to the robot, and orientation across trials, enforcing a minimum deviation of $3$\,cm in translation and $10$$^\circ$ in rotation (along at least one axis) from the previous configuration. Furthermore, we vary lighting conditions and scene layouts, and introduce visual distractions, as illustrated in Fig.~\ref{fig:generalizability}.
Tables~\ref{tab:object-pose} and \ref{tab:success-summary} show that \Method{} achieves the highest success rate and Avg. Len. on all five tasks. The evaluated YOTO variant achieves lower success rates on tasks requiring precise finger placement and contact regulation, particularly in \texttt{Pull Drawer}, \texttt{Open Lid}, and \texttt{Manipulate Pipette}. 
On \texttt{Manipulate Pipette}, pickup succeeds in $28/33$ trials with \Method{}
and $26/33$ with YOTO, whereas the first button press succeeds in $23/33$ and
$16/33$ trials, respectively.
The larger gap at actuation highlights the importance of evaluating interactions beyond initial grasp acquisition.
Compared with OKAMI, direct egocentric wrist-hand extraction and tactile regulation reduce sensitivity to reconstruction errors and contact mismatch. The $\pi_{0.5}$ baseline remains challenged by the humanoid's embodiment and high-DoF action interface despite task-specific finetuning. 
Removing tactile feedback reduces mean success from $76.97\%$ to $63.64\%$,
with the largest decrease on \texttt{Manipulate Pipette}
($63.64\%$ to $42.42\%$).

\begin{figure}[!htb]
    \centering
    \includegraphics[width=\columnwidth,pagebox=cropbox]{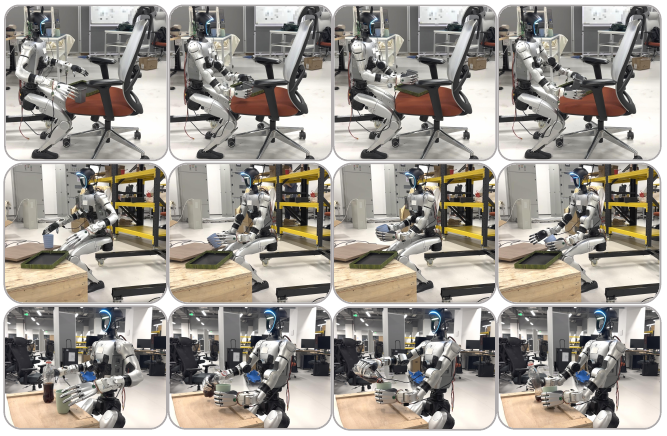}
    \vspace{-0.2in}
    \caption{Cross-object reuse from one demonstration per task. Top and middle: \texttt{Hand Over} transfers a paper-cup demonstration to a toy bin and a plastic cup, respectively. Bottom: \texttt{Pour Water} transfers a bottle-and-paper-cup demonstration to a transparent bottle and a plastic cup.}
    \label{fig:scene-workspaces}
\end{figure}

\Method{} achieves a $90.91\%$ success rate on \texttt{Hand Over} and \texttt{Pour Water}. Other tasks impose tighter contact constraints: \texttt{Pull Drawer} requires grasping a small handle, \texttt{Open Lid} involves a deformable ribbon, and \texttt{Manipulate Pipette} requires a rack-constrained grasp that permits button actuation. Observed failures include collisions while approaching the first keyframe, finger-handle disengagement, ribbon grasp failures, and accidental contact with the pipette's adjacent tip-ejection button. 
\Method{} also supports cross-object skill generalization by reusing each task's original demonstration with a toy bin,
plastic cup, or cube for \texttt{Hand Over}, and with a
transparent bottle and plastic cup for \texttt{Pour Water}.
Fig.~\ref{fig:scene-workspaces} shows representative examples.

{\clubpenalty=10000 \widowpenalty=10000
\textbf{\textit{(Q3)} Robustness to external interference across all stages.} From a $3$\,m starting distance, we perturb \texttt{Pour Water} execution during navigation and calibration together, calibration alone, or manipulation. Table~\ref{tab:recovery} reports success rates of $70\%$, $90\%$, and $90\%$, respectively. Updated navigation and object poses enable re-approach and posture correction, while manipulation replanning supports recovery from interaction failures. Upon failure, Praxis re-estimates object poses and restarts from the first keyframe with updated EE targets, allowing up to three retries per trial. 
Table~\ref{tab:vln-calib-ablation} evaluates the contribution of posture calibration under lateral interference from a fixed initial pose $3$\,m from the workspace. 
Under lateral interference, posture calibration raises preparation-stage completion from $20\%$ to $80\%$ and task completion from $10\%$ to $70\%$, while average navigation inference counts remain similar ($13.2$ versus $13.5$).\par}

\section{Conclusion}\label{main:con}

In this paper, we propose \Method{}, a hierarchical framework for long-horizon humanoid whole-body manipulation tasks. \Method{} coordinates three stages: navigation to the manipulation region, posture calibration for a reachable whole-body configuration, and dexterous manipulation through object-centric physical interaction priors. By re-grounding 6D wrist poses and fine-grained finger motions with online perception, \Method{} transfers human demonstrations to new object poses, scenes, and task conditions without task-specific manipulation-policy retraining. 
Experiments on diverse whole-body manipulation tasks demonstrate strong generalization across spatial, visual, and object variations, as well as robustness to physical disturbances across all three stages. Future work includes bridging the gap between G1-collected demonstrations and unconstrained internet videos or generated videos to further scale up manipulation skills, and integrating higher-level autonomy modules, such as continual learning within the operational space, to better interpret and accomplish user-specified tasks.

\bibliographystyle{IEEEtran}
\bibliography{references}

\appendix

\subsection{Comparison between Recent Whole-body Manipulation Systems and \Method{}}\label{appendix:related-works1}

\begin{table*}[ht]
\centering
\caption{Comparison between recent whole-body manipulation systems and \Method{}. Vis. \& Lang. Input indicates vision--language input, and Long Dist. refers to long-distance navigation.}
\label{tab:loco_comparison}
\renewcommand{\arraystretch}{1.12}
\setlength{\tabcolsep}{7pt}
\begin{tabular}{lcccc}
\toprule
Method
& Dexterity
& One-shot
& Learn from Video
& Vis. \& Lang. Input \\

\midrule

WholeBodyVLA \cite{jiang2025wholebodyvla}
& \xmark & \xmark & \cmark & \cmark \\

HMC \cite{wei2025hmc}
& \xmark & \xmark & \xmark & \xmark \\

DemoHLM \cite{fu2025demohlm}
& \xmark & \cmark & \xmark & \xmark \\

TrajBooster \cite{liu2025trajbooster}
& \cmark & \xmark & \xmark & \cmark \\

OKAMI \cite{li2024okami}
& \cmark & \cmark & \cmark & \xmark \\

\midrule

\textbf{\Method{} (Ours)}
& \cmark & \cmark & \cmark & \cmark \\

\bottomrule
\end{tabular}

\vspace{0.5em}

\begin{tabular}{lccc}
\toprule
Method
& Long Dist. ($\ge$ 5\,m)
& Long Horizon
& Recovery \\

\midrule

WholeBodyVLA \cite{jiang2025wholebodyvla}
& \cmark & \xmark & \cmark \\

HMC \cite{wei2025hmc}
& \xmark & \xmark & \xmark \\

DemoHLM \cite{fu2025demohlm}
& \xmark & \xmark & \xmark \\

TrajBooster \cite{liu2025trajbooster}
& \xmark & \xmark & \xmark \\

OKAMI \cite{li2024okami}
& \xmark & \xmark & \xmark \\

\midrule

\textbf{\Method{} (Ours)}
& \cmark & \cmark & \cmark \\

\bottomrule
\end{tabular}
\vspace{-0.08in}
\end{table*}

As shown in Appendix Tab.~\ref{tab:loco_comparison}, we compare four recent whole-body manipulation systems along eight dimensions—dexterity, one-shot, learning from video, vision-and-language input, long-distance navigation ($\ge$ $5$ m), long-horizon execution, recovery, and training-free deployment. Here, long-distance navigation indicates that the minimum distance the robot must travel to reach the manipulation workspace is at least $5$\,m, and long-horizon execution means the manipulation itself contains at least four distinct steps. WholeBodyVLA \cite{jiang2025wholebodyvla} leverages large-scale action-free videos and supports vision-language conditioning with broad spatial coverage, but it remains limited in fine dexterous manipulation and long-horizon task organization. TrajBooster \cite{liu2025trajbooster} improves cross-embodiment transfer and whole-body coordination via trajectory lifting and subsequent adaptation, yet it still relies on training and a small amount of real data. DemoHLM \cite{fu2025demohlm} can scale from a single demonstration by synthesizing large amounts of simulation data and training policies, but it is not equivalent to real-video one-shot learning and still requires training. HMC \cite{wei2025hmc} focuses on robust contact-rich control through multi-mode control switching, addressing low-level execution rather than closed-loop capabilities such as language-guided navigation, long-distance mobility, and training-free task deployment. In contrast, \Method{} unifies navigation and manipulation in a single pipeline: it uses vision-language navigation and calibration policy to reach targets over long distances, enables training-free deployment to new tasks via one-shot human video demonstrations, and incorporates a recovery mechanism (failure detection and replanning) to support long-horizon, multi-stage execution.

\subsection{Details on Hardware System}\label{appendix:hardware}

\subsubsection{Robot Embodiment} 
Humanoid platforms extend the workspace of bimanual manipulation beyond fixed-base arms and are suitable for whole-body manipulation inspired by human egocentric videos.
We adopt the Unitree G1 EDU humanoid robot\footnote{https://www.unitree.com/g1}, with $27$ degrees of freedom (DoFs). We establish a wireless TCP interface between the robot and a host desktop equipped with an NVIDIA RTX 4090 GPU. This architecture supports real-time, asynchronous perception, planning, and control over a shared Wi-Fi network.

\subsubsection{End-Effector Selection} 
Since the stock G1 humanoid's rubber hands do not support dexterous manipulation, we replace them with two BrainCo Revo2 hands\footnote{https://www.brainco.cn/en-US/products/revo2}, where each is a five-fingered tactile-sensing hand with $11$ degrees of freedom (DoFs), with three in the thumb and two in each of the index, middle, ring, and pinky fingers. 
Each of the two Revo2 hands is connected to the G1 via an RS485-to-Type-C interface operating at a baud rate of 1M, enabling high-speed communication. The integrated tactile sensing improves the hand’s capability for precise and robust manipulation, particularly for articulated and deformable object interactions, as demonstrated in our real-world experiments.

\subsubsection{Camera Observation}
The G1 humanoid is originally equipped with an Intel RealSense D435i RGB-D camera embedded\footnote{https://www.realsenseai.com/products/depth-camera-d435i} integrated into the head with a fixed downward pitch of $48^\circ$, which provides a fixed and limited field of view (FoV). To support long-range perception for navigation, we additionally mount a RealSense L515 solid-state time-of-flight LiDAR camera\footnote{https://www.realsenseai.com/products/realsense-lidar-camera-l515} on the forward-facing surface of the torso. 
The L515 is used for navigation perception, while the D435i supports pre-manipulation calibration and manipulation.
The D435i operates at a resolution of $1280\times720$ at $25$ Hz, with depth frames aligned to the RGB stream, while L515 operates at a resolution of $640\times480$ at $25$ Hz. 

Fig.~\ref{fig:hardware} illustrates our hardware configuration, where the bracketed text denotes how each device is connected to the G1 onboard computer. To ensure responsive control, all I/O, including dual-camera image acquisition and proprioception, is handled asynchronously, avoiding stalls in the main control thread and enabling the whole-body controller to run reliably at $50$Hz.

\subsection{Details on Dataset Construction for Fine-tuning NaVILA}\label{appendix:fine-tune-navila}

To adapt the general-purpose VLN model to our robotic embodiment and experimental environment, we construct a dedicated fine-tuning dataset. Beyond the strategy described in the main paper, we detail our model initialization, object-diverse scene augmentation, and evaluation protocol, and multi-faceted randomization strategy in data collection.

\subsubsection{Model Initialization and Action Space}
We initialize our policy using the official NaVILA checkpoint. Crucially, we select the version that has already undergone extensive fine-tuning on large-scale navigation datasets. We leverage this established locomotor before adapting the model for coarse-level navigation in our specific environment. The objective of this fine-tuning is to enable the robot to recognize target objects and reach their immediate vicinity, after which a calibration module takes over to achieve manipulation-ready postures.

While the original NaVILA model primarily focuses on longitudinal and rotational movements, we expanded the action space by introducing two additional primitives: $\mathrm{Move\ Left}$ and $\mathrm{Move\ Right}$. This augmentation allows the policy to utilize the robot's holonomic capabilities for a more flexible approach to trajectories during the navigation phase, ensuring efficient traversal without the constraints of non-holonomic kinematics.

Our fine-tuning dataset consists of 
$150$ expert trajectories collected in the real world via teleoperation using a handheld remote controller. The distribution of these trajectories is detailed in Appendix Tab.~\ref{tab:action_dist}.

\renewcommand{\arraystretch}{1.2}
\setlength{\tabcolsep}{5pt} 
\begin{table}[h]
    \centering
    \caption{Distribution of trajectories in the fine-tuning dataset.}
    \label{tab:action_dist}
    \begin{small}
    \begin{tabular}{lc}
        \toprule
        \textbf{Action Type} & \textbf{Count} \\
        \midrule
        Move Forward & $46$ \\
        Turn Right   & $32$ \\
        Turn Left    & $32$ \\
        Move Right   & $20$ \\
        Move Left    & $20$ \\
        \midrule
        \textbf{Total} & $\mathbf{150}$ \\
        \bottomrule
    \end{tabular}
    \end{small}
\end{table}

\subsubsection{Object Diversity and OOD Evaluation Protocol}
To rigorously assess visual robustness, we adopt a strict out-of-distribution (OOD) protocol over scene objects. We curate seven target objects (one shelf and six tables) with high visual diversity in geometry (rectangular vs. circular), appearance (wood grain vs. metal), and size. We enforce a strict train–test split: trajectories for only four table variants are used for training, while the remaining two tables and the shelf are reserved exclusively for testing. This design ensures that our evaluation measures genuine generalization to unseen objects and environments rather than memorization.

\subsubsection{Geometric Randomization}
We vary the robot's initial pose, both radial distance and approach angle relative to the target, to encourage viewpoint invariance.

\subsubsection{Instruction Augmentation Strategy}
To enhance the robustness of instruction-following, we implement a ``Semantic Consistency, Stylistic Diversity'' annotation strategy. For each trajectory, we generate multiple instruction variants that strictly preserve the core semantic intent (e.g., target object and action type) while varying linguistic structure. 
For instance, a $\mathrm{Move\ Forward}$ action targeting a table is annotated with diverse phrasings such as:
\begin{itemize}
    \item ``Walk ahead and stop at the table.''
    \item ``Proceed forward and stop beside the table.''
\end{itemize}
This strategy prevents the model from overfitting to specific lexical patterns and encourages deep semantic alignment. 

\subsubsection{Scalable Hybrid Data Acquisition}
We combine robot-collected and human-collected trajectories to scale data efficiently. Human operators are instructed to match the robot’s camera height and motion dynamics, preserving domain consistency while expanding coverage across diverse scenes (e.g., different room layouts and table geometries).

\subsection{IK-error Workspace Analysis}
\label{appendix:ik-error}
Appendix Fig.~\ref{fig:ik-error-thick} visualizes the IK error distributions for both arms' EE, accordingly we set $(x_{\mathrm{ref}}, y_{\mathrm{ref}})=(0.4,0)$ where $0.4$ (meter) is the farthest forward position with tolerant IK error, and $0$ lies at the center of the low-error lateral region.


\subsection{Qualitative Results of Wrist Detection and Dexterous Retargeting across Five Tasks}\label{appendix:onw-shot-teaching}
\begin{figure*}[t]
    \centering
    \includegraphics[width=\linewidth,page=4]{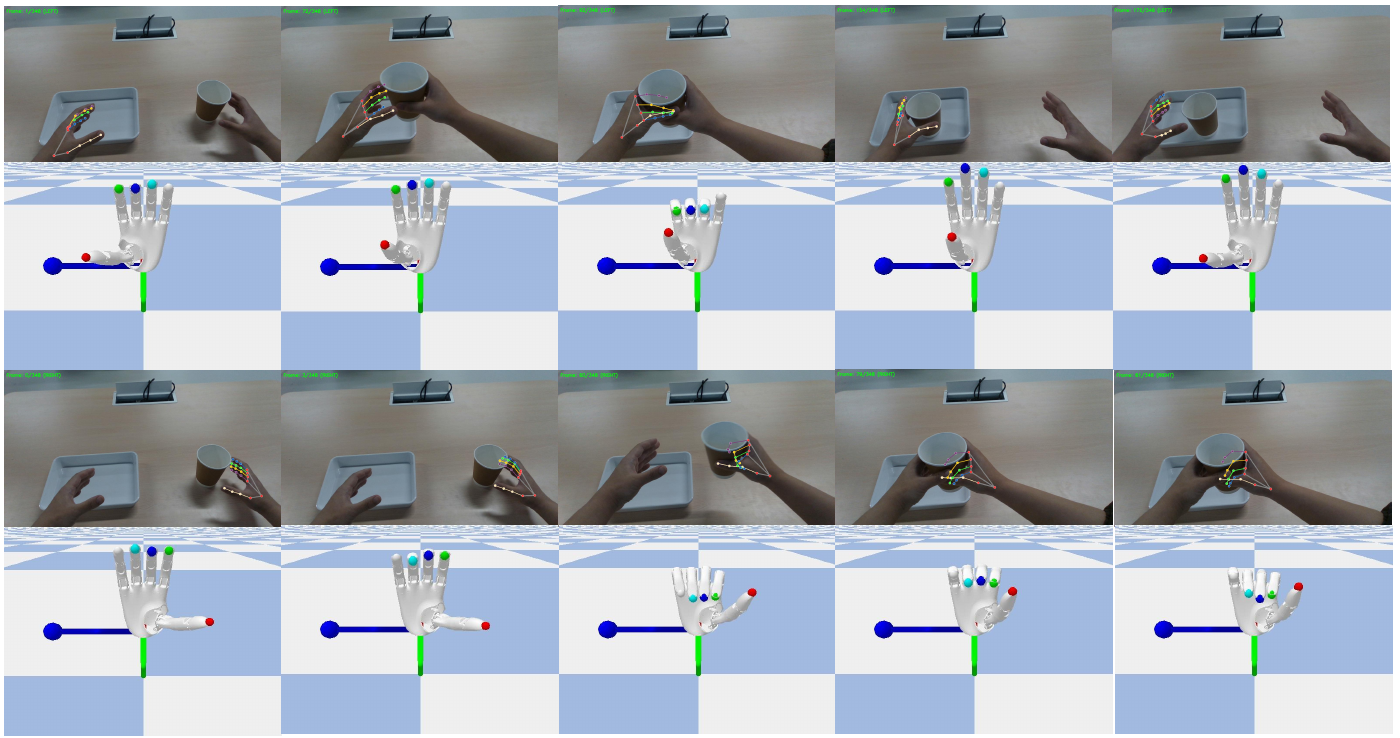}
    \caption{Visualization of wrist pose detection and dexterous retargeting in task \texttt{Hand Over}.}
    \label{fig:hand-over-qualitative}
\end{figure*}

\begin{figure*}[t]
    \centering
    \includegraphics[width=\linewidth,page=5]{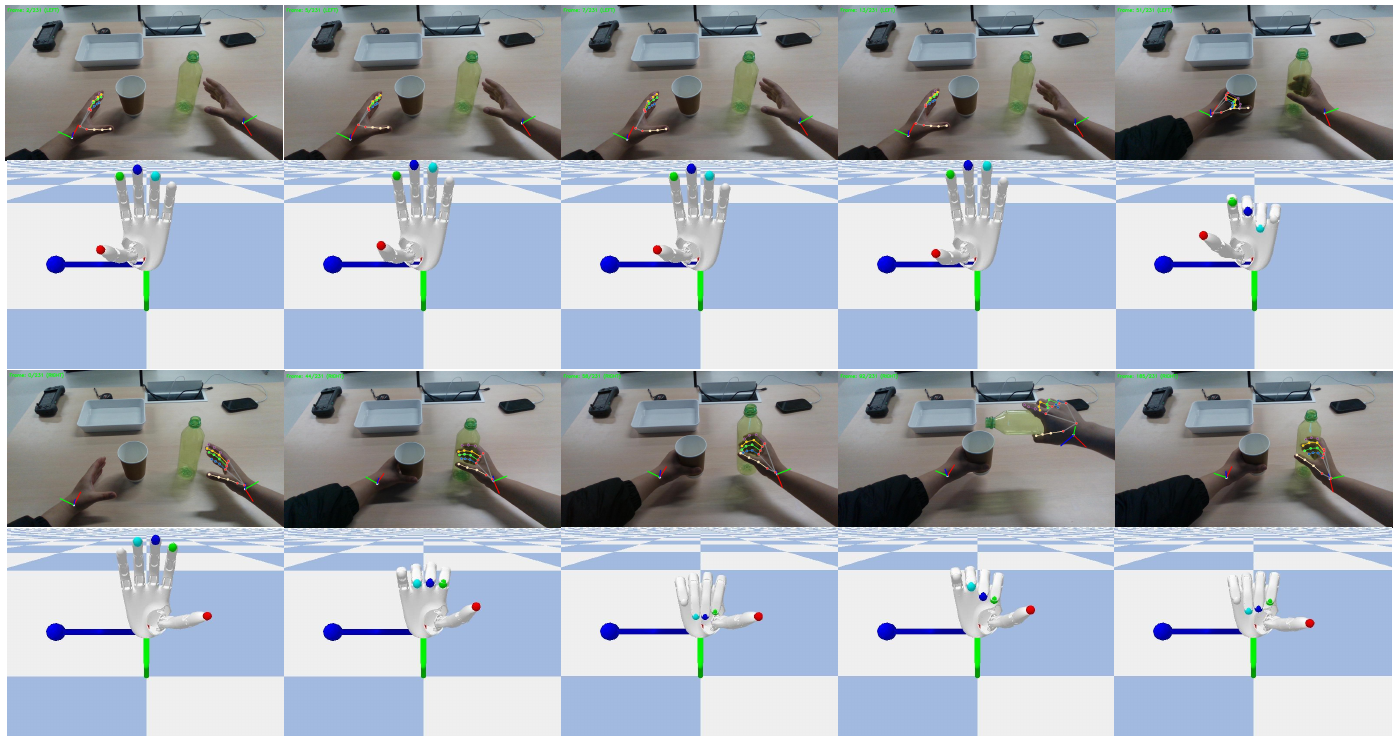}
    \caption{Visualization of wrist pose detection and dexterous retargeting in task \texttt{Pour Water}.}
    \label{fig:pour-water-qualitative}
\end{figure*}

\begin{figure*}[t]
    \centering
    \includegraphics[width=\linewidth,page=1]{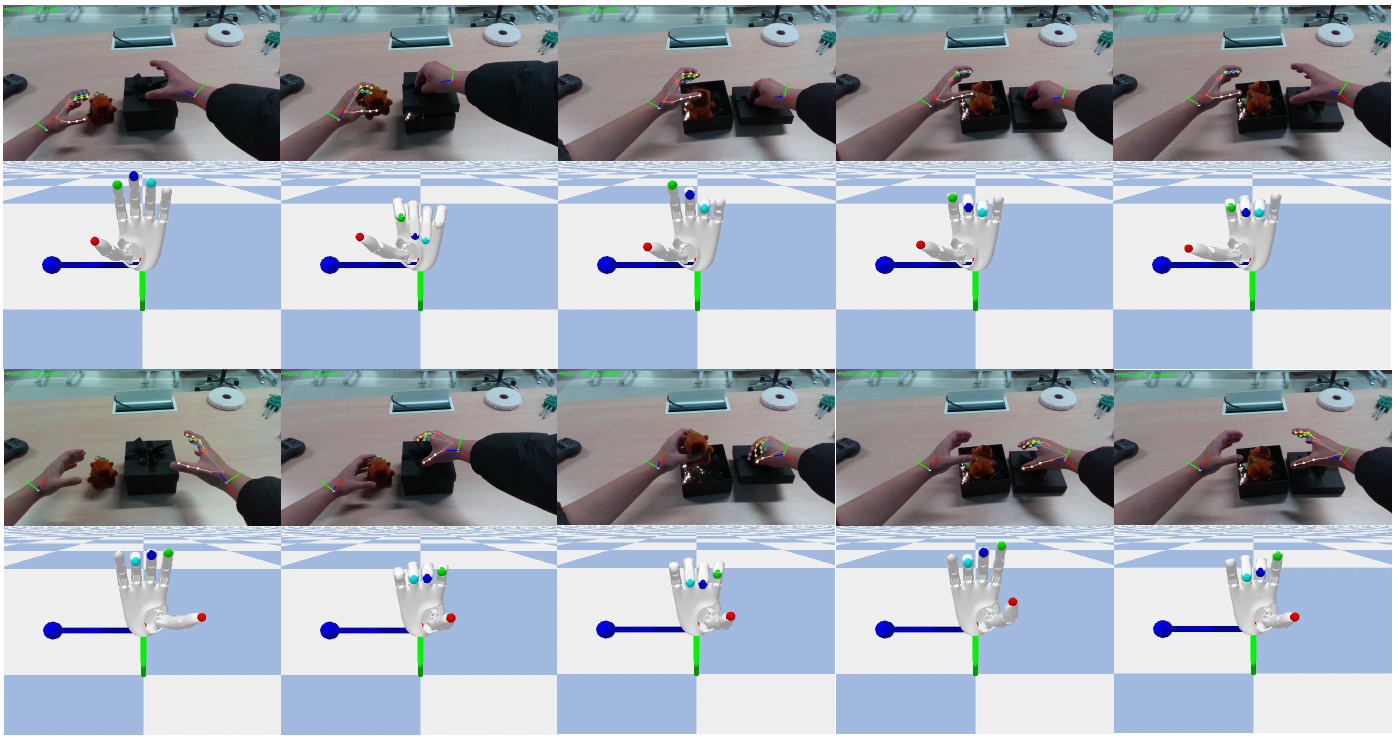}
    \caption{Visualization of wrist pose detection and dexterous retargeting in task \texttt{Open Lid}.}
    \label{fig:open-lid-qualitative}
\end{figure*}

\begin{figure*}[t]
    \centering
    \includegraphics[width=\linewidth,page=2]{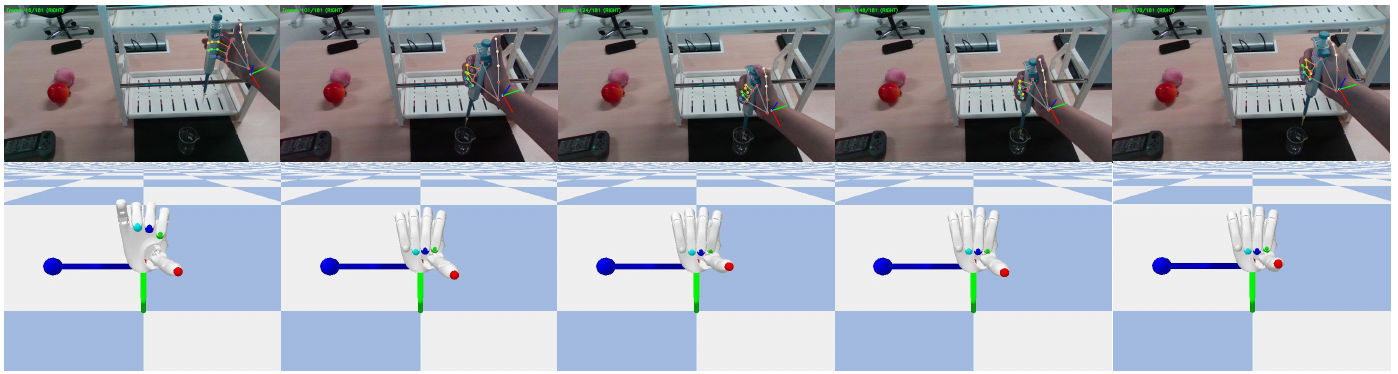}
    \caption{Visualization of wrist pose detection and dexterous retargeting in task \texttt{Manipulate Pipette}.}
    \label{fig:use-pipette-qualitative}
\end{figure*}

Fig.~\ref{fig:pull-drawer-qualitative}, Appendix Figs. \ref{fig:hand-over-qualitative}, \ref{fig:pour-water-qualitative}, \ref{fig:open-lid-qualitative}, and \ref{fig:use-pipette-qualitative} show qualitative results for four bimanual tasks. Each figure visualizes the detected wrist poses for the left and right hands in the first and third rows, respectively, and the corresponding dexterous retargeting results on the Revo2 hand in the second and fourth rows. Appendix Fig.~\ref{fig:use-pipette-qualitative} shows the detected right-wrist poses and the corresponding Revo2 retargeting results for \texttt{Manipulate Pipette}, where only the right hand is used. Across these tasks, the predicted wrist trajectories and retargeted finger motions closely follow the human demonstrations, providing reliable robot-centric supervision for one-shot teaching.

\subsection{Object Masking via LISA for Robust Perception in FoundationPose++}\label{appendix:lisa}
To facilitate accurate 6D pose estimation through FoundationPose++, we utilize LISA (Large Language Instructed Segmentation Assistant) \cite{ravi2024sam} to generate high-fidelity object masks as its input. Unlike contemporary open-vocabulary pipelines such as Grounded-SAM \cite{ren2024groundedsamassemblingopenworld}, which often struggle with implicit instructions or scenarios requiring complex semantic reasoning (e.g., "Find the object that is blocking the door" or "Find the screwdriver needed to fix this toy"), LISA employs an embedding-as-mask paradigm that integrates segmentation directly into a multi-modal LLM's reasoning process by expanding the vocabulary with a \texttt{<SEG>} token.

This integrated approach offers a dual advantage for the \Method{} framework by combining superior linguistic sensitivity with extensive world knowledge to handle specialized tasks. In instances where standard models produce masking errors due to scene complexity or visual ambiguity, LISA allows users to provide more descriptive or context-aware prompts, leveraging its internal reasoning to actively correct and refine the segmentation. This capability is particularly vital for identifying and segmenting specialized or out-of-distribution objects, such as the pipette used in our dexterous manipulation experiments, which are frequently unrecognized by conventional segmentors but are well-represented within the LLM's pre-trained knowledge base.

\subsection{Details and Qualitative Results of Tactile-informed Grasping}\label{appendix:tactile}
To illustrate tactile-informed grasping, we visualize in Appendix Fig. \ref{fig:revo2-tactile} the fingertip normal (blue) and tangential (orange) forces applied to the deformable toy in \texttt{Open Lid} under two-, three-, and five-finger grasps. While we compare these configurations for analysis, all experiments in the main paper use five-finger grasps to prioritize safety and robustness.


\subsection{Detailed Task Descriptions}\label{appendix:taskdescription}
We describe each task in detail, where L, R, and L\&R indicate steps performed by the left hand, right hand, and both hands, respectively.
\begin{itemize}
    \item \texttt{Hand Over:} involves a paper cup and a plastic tray; and consists of $5$ steps: pick cup (R), close to left hand (R), hand over (L\&R), move to tray (L), place cup (L).
    \item \texttt{Pour Water:} involves a bottle with liquid and a paper cup; and consists of $6$ steps: pick cup (L), pick bottle (R), close to each other (L\&R), pour water (R), place bottle (R), place cup (L).
    \item \texttt{Pull Drawer:} involves a drawer with two handles and a building block; and consists of $4$ steps: pull drawer (L), pick block (R), place block (R), push drawer (L).
    \item \texttt{Open Lid:} involves a box with a lid and a toy; and consists of $5$ steps: lift lid (R), pick toy (L), move lid away (R), place toy in box (L), place lid (R).
    \item \texttt{Manipulate Pipette:} involves a pipette, an experiment shelf, and a plastic tray; and consists of $5$ steps: pick pipette (R), press button (R, aspiration), release button (R), press button (R, dispensing), release button (R).
\end{itemize}

\subsection{Details on Teleoperation and Preprocessing for VLA Baselines}\label{appendix:vla-dataset}
The observation space is structured as follows: dimensions $0$--$6$ correspond to the left arm, $7$--$13$ to the right arm, $14$--$19$ to the left hand, and $20$--$25$ to the right hand. We collect $85$ trajectories per task via teleoperation at the same resolution and frame rate, and include them in the finetuning dataset. 
Teleoperation is performed using the Meta Quest 3\footnote{\url{https://www.meta.com/ae/quest/quest-3/}} as the XR (Extended Reality) device. The training batch size is set to 128, and we train for 5000 steps to ensure the VLA baseline can adequately learn the provided trajectories without overfitting.

During the preprocessing phase, we performed frame-level alignment for each episode, retaining only valid frames that contained both image and observation data. Each frame was assigned a timestamp calculated as $t = \text{frame\_index} / 30$, along with a local frame index, an episode index, and a global index. 
The \texttt{next.done} flag was set to true for the final frame of each trajectory, and \texttt{next.reward} was initialized to zero. 

Following preprocessing, we converted the dataset into the LeRobot \cite{cadene2024lerobot} standard formats to support different training requirements: LeRobot v2.1 for GR00T N1.6 and v3.0 for $\pi$ series.

\subsubsection{v2.1 Conversion (Episode-Based)}
We organized the raw data into the LeRobot v2.1 format, utilizing an episode-based structure. Observations and RGB images were read frame-by-frame, temporally aligned, and written into individual Parquet files per episode, where both \texttt{observation.state} and \texttt{action} were stored as $26$-dimensional vectors. Image sequences were encoded as H.264 MP4 videos at $30$ FPS, with one video file generated for each episode. 

Comprehensive metadata was generated, including:
\begin{itemize}
    \item \texttt{meta/info.json}: specifies state and action dimensions, FPS, path rules, and indexing.
    \item \texttt{meta/episodes.jsonl}: records episode lengths.
    \item \texttt{meta/tasks.jsonl}: contains textual task descriptions.
    \item \texttt{meta/episodes\_stats.jsonl} and \texttt{meta/stats\\.json}: stores statistical metrics (min, max, $1^{st}/99^{th}$ quantiles, etc.).
    \item \texttt{meta/modality.json}: defines the mapping between the visual input (head camera) and robot components (arms/hands).
\end{itemize}

\subsubsection{v2.1 $\rightarrow$ v3.0 Conversion (File-Based)}
To facilitate efficient training for the $\pi$ series, we consolidated the v2.1 data into the v3.0 file-based structure. This involved concatenating Parquet rows from multiple episodes into chunked files (e.g., \texttt{data/chunk-*/file-*.parquet}) and merging corresponding video segments into unified video files (e.g., \texttt{videos/\{video\_key\}/chunk-*/file-*.mp4}). 

Metadata was updated accordingly: \texttt{meta/episodes/*.parquet} was generated to index the range and video timestamps of each episode within the merged files, and \texttt{meta/tasks.parquet} was created for the task registry. Finally, \texttt{meta/info.json} was updated with the new path templates and version information. This transformation converts the data from episode-based storage to a highly efficient, streamable v3.0 format without altering the underlying data semantics.

\subsection{Additional Implementation Results in Simulation}
\label{appendix:experiment_simulation}
To augment our real-world findings, we evaluate the keyframe-based motion framework within high-fidelity simulation environments built in NVIDIA Isaac Sim \cite{NVIDIA_Isaac_Sim}, utilizing digital assets from the ArtVIP dataset \cite{jin2025artviparticulateddigitalassets}, a comprehensive library of $206$ high-fidelity articulated objects optimized for robotic manipulation. We leverage a Meta Quest 3 headset for teleoperation to facilitate scalable and efficient expert trajectory collection, during which the motion data undergoes a significant abstraction process: by identifying and processing local motion extrema, we isolate a compact set of approximately $10$ to $20$ keyframes for each trajectory, representing a substantial refinement from the original $500$ to $1000$ frames. Our empirical observations indicate that simple linear interpolation between these sparse keyframes robustly reconstructs the original demonstration trajectories while preserving interaction stability, as illustrated in Appendix Figure \ref{fig:simulation_results}. These retargeted motion priors enable the robot to execute diverse manipulation primitives, confirming the kinematic feasibility of our keyframe-based representation before real-world deployment.

\begin{figure*}[h]
    \centering
    \includegraphics[width=0.9\textwidth]{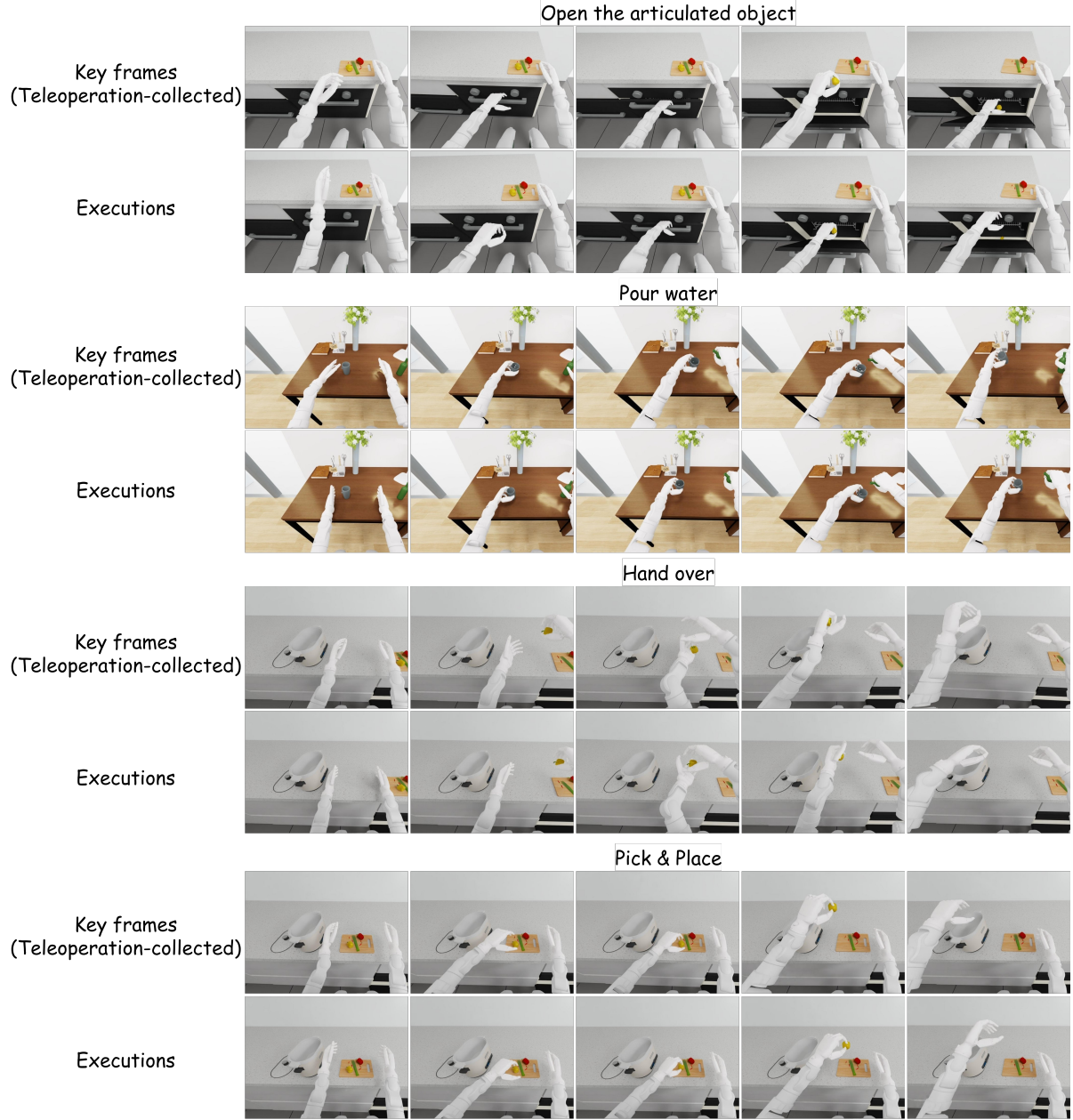}
    \caption{Sequential snapshots of experiments in simulation. The top and bottom rows for each task display teleoperation-collected keyframes and their corresponding execution sequences, respectively. These results demonstrate the efficacy of our keyframe-based representation in maintaining precise interaction geometry across diverse manipulation tasks.}
    \label{fig:simulation_results}
\end{figure*}

\subsection{Qualitative Snapshots of Experimental Process across Five Tasks} \label{appendix:experiment_additional}
This section provides a comprehensive visual walkthrough of the five real-world whole-body manipulation tasks conducted in our empirical studies. As visualized in Appendix Fig. \ref{fig:experiment_additional}, \Method{} demonstrates the ability to coordinate bimanual control and dexterous finger motions to handle objects spanning transparent, rigid, articulated, deformable, and tool-use categories. These sequential snapshots (shadow means earlier state) highlight the system's proficiency in managing long-horizon tasks through a unified pipeline of perception-conditioned execution.

\begin{figure*}[h]
    \centering
    \includegraphics[width=0.85\textwidth]{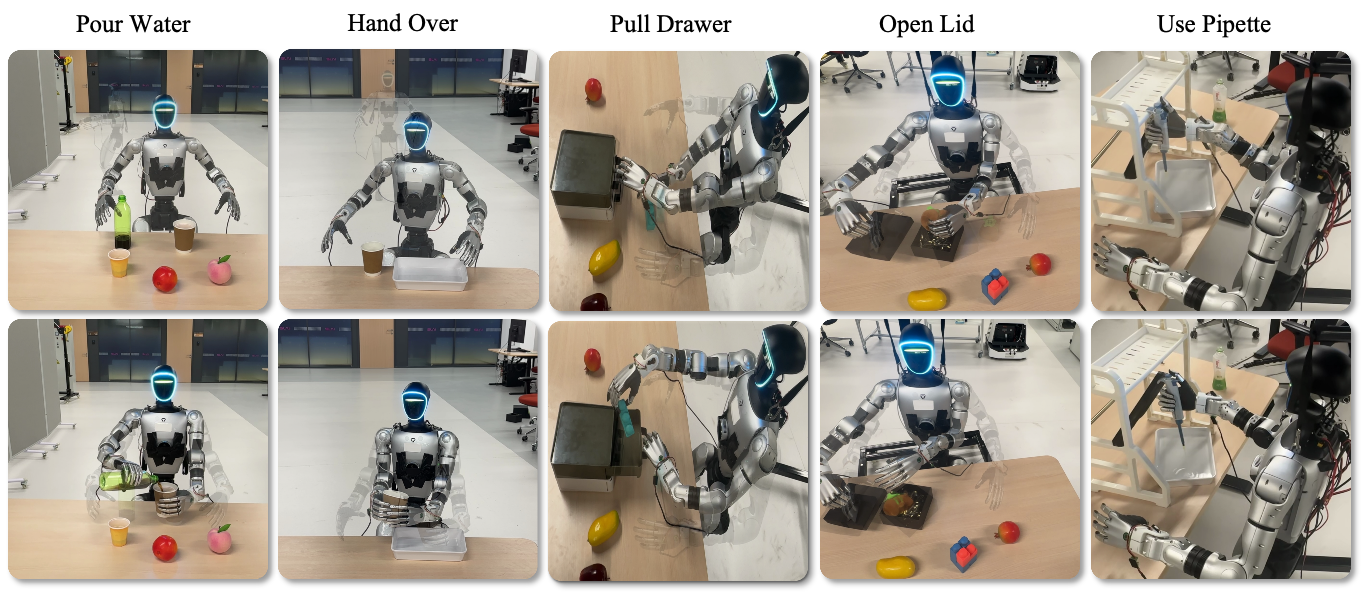}
    \caption{Sequential snapshots showing the progression of the experiment across five tasks. Semi-transparent G1 indicates historical states.}
    \label{fig:experiment_additional}
\end{figure*}

\subsection{Detailed Analysis of Failure Cases}\label{appendix:failure-analysis}
Although \Method{} demonstrates strong performance on long-horizon whole-body manipulation, we observe several recurring failure modes in real-world evaluation. We analyze execution failures across the five tasks that are reported in the main paper.

\subsubsection{Hand Over \& Pour Water} A common failure in these two tasks occurs at the very beginning of the manipulation stage: the target object is sometimes placed along the robot’s transition path from its default posture to the first keyframe. As a result, the robot collides with the object before reaching the initial pose, causing early termination. This explains why the overall pick success rate is not $100\%$, despite the manipulation policy being effective once the first keyframe is reached.

\subsubsection{Pull Drawer} During drawer pulling, failures are often caused by finger–handle disengagement. The handle provides a limited contact area and can induce slip or rolling contact under imperfect approach angles, leading to an unstable grasp and unsuccessful opening. In addition, when lifting an object near the drawer, the carried object may collide with the drawer front or surrounding structure, causing it to drop during transport and reducing the success rate of object placement.

\subsubsection{Open Lid} The gift box lid is primarily grasped via a thin ribbon, which is a challenging target for finger closure. In several cases, the fingers fail to securely capture the ribbon due to its small thickness and deformability, resulting in an unsuccessful lift of the lid.

\subsubsection{Manipulate Pipette} The pipette contains an additional nearby button used to eject the disposable tip. This tip-ejection button is located close to the aspiration button and can be accidentally contacted during execution. Such unintended presses change the hand’s contact configuration, disturb the intended finger placement, and reduce effective force transmission to the aspiration button, leading to failed button presses and contributing to the observed drop in success rate.

\begin{figure*}[h]
    \centering
    \includegraphics[width=0.6\linewidth]{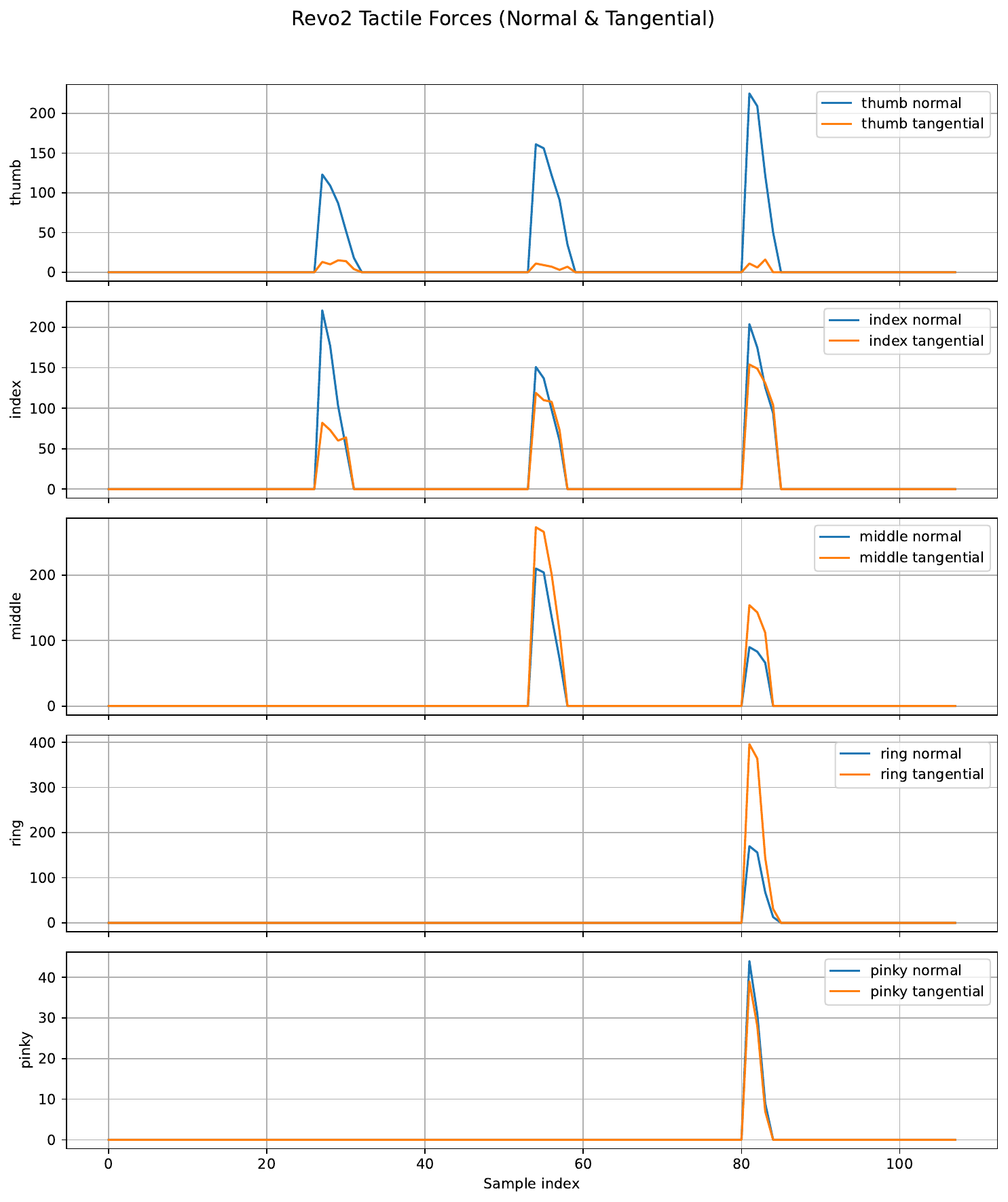}
    \caption{Visualization of fingertip tactile forces when grasping the toy (step \textcolor{RoyalBlue}{pick toy}) in \texttt{Open Lid} using two-, three-, and five-finger configurations of the left Revo2 hand.}
    \label{fig:revo2-tactile}
\end{figure*}

\end{document}